# Reliability-Aware Short-Term Glycemic Risk Prediction Using an Anchor Based Comparative Framework of Sequence, Multimodal, and State-Updating Digital-Twin Models

PIR BAKHSH KHOKHAR, Department of Computer Science, University of Salerno, Italy
CARMINE GRAVINO, Department of Computer Science, University of Salerno, Italy
FABIO PALOMBA, Department of Computer Science, University of Salerno, Italy

Anticipating short-term hypoglycemic and hyperglycemic risk is essential for preemptive diabetes management. However, current machine learning approaches often use varied prediction parameters and evaluation protocols and focus primarily on discriminative performance. This study introduces an anchor-based benchmarking framework for short-term glycemic-risk prediction utilizing longitudinal glucose sequences and multimodal patient information. We converted the T1DiabetesGranada dataset into over 5.5 million standardized prediction anchors from 736 individuals with type 1 diabetes, enabling standardized comparisons of hypoglycemia and hyperglycemia prediction at 30-, 60-, and 120-minute intervals. We evaluated three neural model families—a Hybrid LSTM–Transformer, a Multimodal Digital Twin, and a State-Updating Digital Twin with gap-aware longitudinal state progression—under identical patient-level validation conditions. In addition to predictive performance, the models were evaluated for calibration, threshold behavior, explanation reliability, cross-model disagreement, and robustness to controlled perturbations. Although no statistically significant differences in predictive performance were identified after Holm correction, the models displayed complementary reliability characteristics. The Multimodal Digital Twin attained the lowest mean Brier score, the State-Updating Digital Twin achieved the lowest mean expected calibration error (ECE-10), and the Hybrid LSTM–Transformer exhibited high explanation faithfulness and stability under its temporal attribution protocol, while the digital-twin models displayed distinct model-specific explanation characteristics. These findings indicate that the significance of digital twins extends beyond predictive accuracy and should be assessed using reliability and deployment-oriented criteria. The proposed framework provides a replicable basis for benchmarking longitudinal healthcare AI systems and highlights the importance of assessing reliability and conducting external validation before clinical deployment.

## 1 Introduction

Type 1 diabetes (T1D) management necessitates the timely anticipation of short-term glycemic fluctuations to enable preventive measures before clinically significant hypoglycemia or hyperglycemia occurs. Continuous glucose monitoring (CGM) provides longitudinal measurements of glucose dynamics and clinically relevant indicators, including time in range, time below range, and time above range [5]. These advances have spurred research on glucose forecasting, hypoglycemia prediction, personalized CGM analytics, and data-driven glycemic-risk modeling [9, 17, 24].

Authors' Contact Information: Pir Bakhsh Khokhar, pkhokhar@unisa.it, Department of Computer Science, University of Salerno, Fisciano, Italy; Carmine Gravino, gravino@unisa.it, Department of Computer Science, University of Salerno, Fisciano, Italy; Fabio Palomba, fpalomba@unisa.it, Department of Computer Science, University of Salerno, Fisciano, Italy.

Early research primarily concentrated on predicting future glucose values from CGM trajectories using recurrent neural networks and related temporal models [22, 27, 38]. More recently, attention-based architectures, transformers, state-space models, federated learning, uncertainty-aware learning, and CGM-specific representation models have expanded predictive capabilities [4, 16, 23, 25, 39]. However, increased architectural complexity does not necessarily translate into greater clinical utility. Successful healthcare deployment also requires reliable calibration, robustness, reproducibility, transparent reporting, and evaluation under realistic clinical conditions [10, 20, 42, 44].

Despite this progress, three significant limitations persist. First, many studies focus on single-value glucose forecasting, isolated glycemic events, or individual prediction horizons, limiting evidence for unified prediction of both hypoglycemic and hyperglycemic risk across multiple horizons. Second, most approaches primarily rely on recent glucose trajectories while underutilizing patient-specific information, including biochemical measurements, diagnostic history, monitoring density, missingness, and temporal-gap patterns. Although multimodal representation learning can integrate heterogeneous healthcare information, its application to longitudinal glycemic-risk prediction remains limited [1]. Third, evaluation often relies on discrimination metrics, despite the importance of reliable probability estimates, appropriate operating thresholds, robust explanations, and systematic assessment under controlled perturbations [8, 10, 13, 42].

Operational data-driven digital twins offer a promising framework for addressing these challenges by maintaining patient-specific computational representations that integrate longitudinal observations with contextual information. Researchers have proposed healthcare digital twins to support personalized monitoring, prediction, and decision-making [7, 11]. Recent diabetes studies have explored personalized data augmentation, physiologically constrained neural representations, hybrid sequence forecasting, and counterfactual intervention modeling [28, 32, 35, 41]. Nevertheless, evidence remains limited regarding whether digital-twin formulations provide measurable reliability or deployment-oriented advantages over strong sequence-based predictors under a common benchmarking protocol.

Fairly comparing glycemic prediction models remains challenging because published studies differ in datasets, preprocessing strategies, prediction definitions, observation windows, and validation protocols. The T1DiabetesGranada dataset provides longitudinal glucose measurements together with demographic, biochemical, and diagnostic information [34], while AZT1D, BrisT1D, and Glucose-ML offer complementary real-world diabetes datasets [18, 21, 33].

To address these challenges, this study transforms the T1DiabetesGranada dataset into a standardized anchor-based representation. Each prediction anchor consists of a six-hour historical glucose window sampled at 15-minute intervals and is linked to six binary outcomes representing hypoglycemic and hyperglycemic risk within 30-, 60-, and 120-minute prediction horizons. This formulation enables comparison of heterogeneous model families under identical preprocessing and patient-level validation conditions.

Beyond predictive performance, reliable healthcare AI requires evaluation of calibration, explanation quality, and model behavior. Explainable AI methods have been applied to glycemic prediction [8, 13, 30], while healthcare AI research emphasizes assessing explanation faithfulness, stability, and consistency rather than visual plausibility alone [2, 10, 40].

This study introduces an anchor-based comparative framework for short-term glycemic-risk prediction using longitudinal glucose trajectories and multimodal patient information. Three neural model families are evaluated under identical anchor construction, preprocessing, prediction targets, and patient-level validation: a Hybrid LSTM–Transformer, a Multimodal Digital Twin, and a State-Updating Digital Twin. The main contributions are:

> Short-term glycemic-risk prediction is formulated as a unified six-task benchmark for hypoglycemia and hyperglycemia prediction at 30-, 60-, and 120-minute horizons using standardized anchor-based CGM and multimodal patient information.

> An operational State-Updating Digital Twin is developed by integrating multimodal fusion, recurrent latent-state propagation, gap-dependent decay, and validity-aware masking for longitudinal patient-state representation.

> A reliability-aware comparison of sequence and digital-twin models is conducted using predictive performance, calibration, explanation analysis, cross-model disagreement, error stratification, and controlled perturbation evaluation under a common patient-level validation protocol.

Rather than asserting universal superiority of digital-twin modeling, this study examines whether multimodal integration and longitudinal state updating offer measurable advantages in predictive reliability and deployment-oriented behavior under standardized healthcare prediction conditions.

The paper is organized as follows. Section 2 reviews related work. Section 3 presents the methodology, model architectures, and evaluation protocol. Section 4 reports the experimental results. Section 5 discusses implications and limitations, and Section 6 concludes the paper.

## 2 Related Work

Prior research on glycemic prediction comprises CGM-based forecasting, adverse-event prediction, temporal deep learning, reliability evaluation, explainability, and digital twins for diabetes. This section analyzes these areas, focusing on prediction formulation, input modalities, validation techniques, and deployment-oriented evaluation.

The growing availability of continuous glucose monitoring (CGM) data has facilitated extensive research on glucose forecasting, hypoglycemia prediction, personalized risk estimation, and data-driven diabetes management [5, 9, 17, 24]. Early studies primarily concentrated on forecasting future glucose values from CGM trajectories using recurrent neural networks and related temporal models [22, 26, 27]. Subsequent research extended this direction to adverse glycemic-event prediction, particularly hypoglycemia, by investigating early-warning performance and generalization across patient populations [15, 38]. However, the reviewed approaches often focus on CGM-based inputs, individual outcomes, or selected prediction horizons, providing limited evidence for unified multimodal prediction of both hypoglycemic and hyperglycemic risk.

> **Current Limitation:** Existing CGM-based prediction studies provide limited evidence for unified multi-horizon prediction of both hypoglycemia and hyperglycemia using multimodal longitudinal patient information.

Recent glycemic prediction methods have evolved toward attention-based architectures, recurrent–transformer hybrids, transformers, state-space models, federated learning, uncertainty-aware learning, and CGM-specific foundation models [4, 6, 12, 16, 19, 23, 25, 29, 36, 39]. These approaches enhance the capacity to model short-term glucose fluctuations and temporal dependencies. However, existing studies differ considerably in datasets, prediction horizons, preprocessing pipelines, personalization strategies, and validation protocols, complicating direct comparison [9, 24]. Architectural complexity alone does not establish clinical utility, particularly when evaluation is limited to forecasting or discrimination metrics [20, 44].

> **Current Limitation:** Advanced glycemic prediction models are commonly evaluated under heterogeneous datasets, preprocessing procedures, prediction horizons, and validation protocols, limiting reliable comparisons across different model architectures.

Reliable healthcare AI requires more than predictive discrimination. Although AUROC and forecasting error are widely reported [14], these metrics alone may not adequately reflect clinical utility under class imbalance [37]. Models with similar discrimination performance can differ in precision, recall, calibration, and alert behavior. Consequently, calibration analysis, threshold evaluation, and reliability-oriented assessment are important for trustworthy healthcare AI systems [10, 42, 43].

Explainability has also received increasing attention in glycemic prediction. Existing studies have applied feature attribution, graph-based explanations, and interpretable machine-learning methods to identify glucose patterns and contextual factors associated with predictions [8, 13, 30]. However, explanations should be assessed beyond visual plausibility by considering faithfulness, stability, transparency, and consistency with model behavior [2, 10, 40]. Evidence regarding these properties, particularly under controlled input perturbations, remains limited in glycemic-risk prediction.

> **Current Limitation:** Existing glycemic prediction studies provide limited evaluation of reliability characteristics beyond predictive performance, including probability calibration, explanation faithfulness, stability, and robustness under controlled perturbations.

Reproducible benchmarking is challenging: diabetes datasets vary in patient populations, available modalities, prediction definitions, and validation protocols. The T1DiabetesGranada dataset provides longitudinal glucose measurements along with demographic, biochemical, and diagnostic information from 736 individuals with T1D [34]. Other resources, such as AZT1D, BrisT1D, and Glucose-ML, offer complementary real-world diabetes data [18, 21, 33]. However, dataset availability alone does not ensure fair comparison, as prediction units, retrospective windows, outcome definitions, preprocessing strategies, and patient-level validation protocols vary across studies [31].

Multimodal representation learning can integrate complementary clinical information into richer patient representations, but effectively using longitudinal multimodal data requires consistent temporal alignment, missing-data handling, and standardized evaluation [1]. An anchor-based formulation provides a common computational unit by defining consistent observation windows, prediction horizons, and patient-level validation boundaries.

Direct comparison of glycemic prediction models remains difficult because existing studies often employ incompatible prediction formulations, preprocessing pipelines, and evaluation strategies.

Supplementary Table S1 provides a detailed comparison of representative studies, including datasets, prediction objectives, input modalities, digital-twin utilization, and evaluation strategies.

Researchers increasingly explore digital twins in healthcare as patient-specific computational representations that integrate physiological, clinical, and longitudinal information to support personalized monitoring, prediction, and decision-making [7, 11]. In diabetes research, digital-twin approaches include physiological simulation, personalized data augmentation, constrained neural architectures, hybrid sequence forecasting, and counterfactual intervention modeling.

Recent studies have explored personalized data augmentation [32], physiologically constrained neural modeling [35], long-horizon glucose forecasting [41], and patient-specific intervention optimization [3, 28]. While these studies highlight the potential of digital twins for diabetes management, their objectives and evaluation settings vary significantly. Evidence remains limited regarding whether digital-twin formulations provide measurable reliability advantages over sequence-based predictors under a common evaluation protocol.

Unlike approaches focused on physiological simulation, data augmentation, or intervention-oriented applications, this work assesses digital twins as reliability-aware predictive representations for short-term glycemic-risk assessment. The comparison focuses on whether multimodal fusion and longitudinal state updating offer measurable differences in calibration, explanation behavior, model agreement, and robustness.

**Current Limitation:** Evidence remains limited regarding whether multimodal and state-updating digital twins enhance reliability characteristics, including calibration, explanation behavior, model agreement, and robustness, compared with strong sequence-based predictors.

### Position of This Study

This study addresses these limitations through a controlled, anchor-based comparative framework that evaluates three model families: a Hybrid LSTM–Transformer, a Multimodal Digital Twin, and a State-Updating Digital Twin. All models are evaluated under identical anchor construction, six-hour observation windows, prediction targets, preprocessing, and patient-level validation.

The proposed framework jointly predicts hypoglycemia and hyperglycemia risk at 30-, 60-, and 120-minute horizons while extending evaluation beyond discrimination metrics to include calibration, threshold behavior, explanation reliability, cross-model disagreement, error stratification, and controlled perturbation analysis. This enables a systematic assessment of whether multimodal fusion and longitudinal state updating provide measurable reliability and deployment-oriented advantages over sequence-based predictors.

## 3 Materials and Methods

The empirical evaluation was structured around three research questions addressing four interrelated limitations identified in previous glycemic-prediction research: limited unified multi-horizon multimodal prediction, heterogeneous evaluation protocols, insufficient controlled evidence regarding multimodal and state-updating digital-twin models, and evaluation focused primarily on predictive discrimination. These limitations are addressed collectively rather than through a one-to-one correspondence between limitations and research questions.

**RQ1** How can heterogeneous longitudinal glucose, static, biochemical, diagnostic, missingness, and temporal-gap information be transformed into a unified anchor-based representation for jointly predicting hypoglycemia and hyperglycemia within 30-, 60-, and 120-minute horizons?

RQ1 addresses the lack of standardized multi-horizon formulations by defining a common six-task anchor-based representation with aligned inputs, retrospective windows, outcome definitions, and patient-level validation boundaries.

**RQ2** Under a common anchor-based, patient-level evaluation protocol, do multitask multimodal fusion and longitudinal state updating offer measurable predictive advantages over strong target-specific sequence baselines?

RQ2 addresses the limited controlled evidence on digital-twin models by comparing multimodal and state-updating formulations with a strong sequence baseline under identical data construction, preprocessing, prediction targets, and patient-level folds.

| RQ3 | How do the compared model families differ beyond predictive discrimination with respect to calibration, operating thresholds, explanation faithfulness and stability, cross-model disagreement, subgroup error patterns, and responses to controlled perturbations? |
| --- | --- |

RQ3 broadens the focus beyond discrimination by evaluating calibration, threshold behavior, explanation reliability, cross-model disagreement, subgroup error patterns, and sensitivity to controlled perturbations.

Together, RQ1 addresses unified multimodal task formulation and standardized evaluation boundaries, RQ2 examines the comparative predictive performance of sequence-based and digital-twin models under a common protocol, and RQ3 investigates reliability characteristics beyond predictive discrimination. Thus, the four identified research limitations are addressed through three complementary research questions.

### 3.1 Overview of the Proposed Anchor-Based Comparative Framework

The proposed framework enables a standardized comparison of sequence-based and digital-twin models for short-term glycemic-risk prediction using longitudinal glucose trajectories and multimodal patient information. As illustrated in Fig. 1, the framework comprises four main components: (i) data preparation and anchor-based task construction, (ii) multimodal patient representation, (iii) sequence-based and digital-twin model development, and (iv) reliability-aware evaluation. These components support the prediction of hypoglycemia and hyperglycemia at 30-, 60-, and 120-minute horizons under a common patient-level validation protocol. The framework is designed for retrospective risk assessment and is not intended for autonomous insulin dosing, treatment recommendation, or direct clinical intervention. The following subsections describe the dataset, input representations, model architectures, training and validation procedures, and evaluation methods.

### 3.2 Dataset and Anchor-Based Task Formulation

The study used the T1DiabetesGranada dataset, comprising longitudinal glucose measurements and multimodal clinical information from 736 individuals with type 1 diabetes [34]. Glucose, biochemical, and diagnostic records were linked using patient identifiers and chronologically aligned. Missing modality information was preserved through explicit availability indicators rather than excluding incomplete patient records.

The harmonized data were transformed into standardized prediction anchors, each containing a six-hour retrospective glucose window, aligned structured variables, and six binary future-risk outcomes. Candidate anchors were generated at four 15-minute steps, corresponding to a nominal one-hour stride. Anchors were retained when at least five glucose measurements were available within the retrospective window and a glucose observation was present at the anchor timestamp. This process produced 5,503,441 prediction anchors from 736 patients.

The six prediction targets represented hypoglycemia and hyperglycemia within 30-, 60-, and 120-minute horizons. For each anchor, future glucose observations within the corresponding horizon were examined. Hypoglycemia was defined as at least one future glucose measurement below 70 mg/dL, whereas hyperglycemia was defined as at least one measurement above 180 mg/dL [5]. Sequence-based models predicted each outcome separately, while the digital-twin models jointly estimated all six risks.

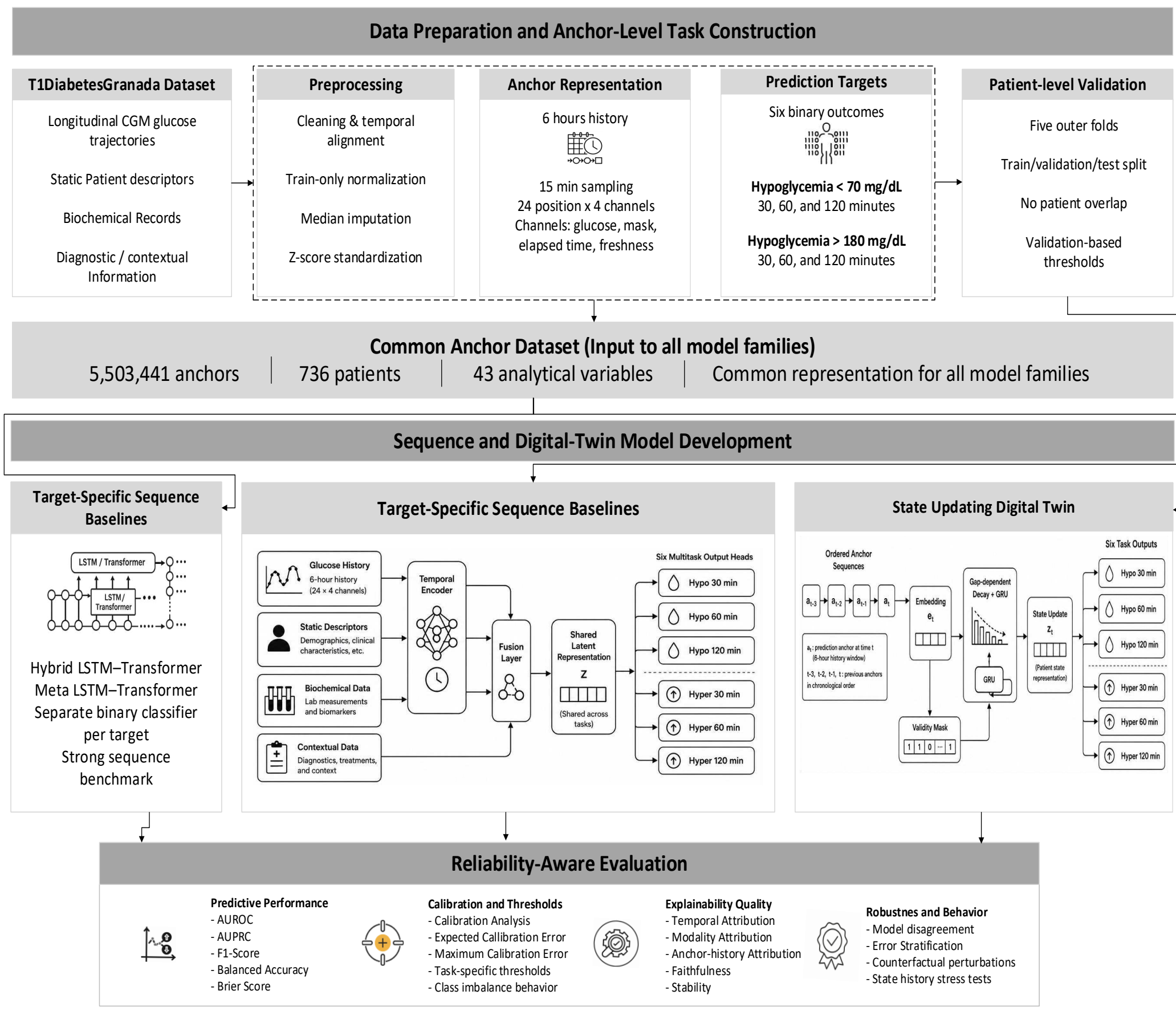

Fig. 1. Overview of the anchor-based comparative framework. Longitudinal glucose trajectories and multi-modal patient descriptors are transformed into standardized anchors and evaluated using sequence baselines, a multimodal digital twin, and a state-updating digital twin.

Because prediction anchors represented repeated longitudinal observations rather than independent samples, all training, validation, and testing partitions were defined at the patient level. All anchors belonging to the same patient were assigned to a single outer fold, preventing overlap of patient histories and future observation windows across development and held-out test partitions.

### 3.3 Input Modalities and Preprocessing

For each anchor, we sampled a six-hour retrospective glucose window at 15-minute resolution, yielding 24 time steps. Each position contained four channels: glucose value, observation mask, elapsed time, and freshness indicator.

$$X^{\text{seq}} \in \mathbb{R}^{24\times4}. \tag{1}$$

Missingness and temporal irregularity were retained as input information rather than removed. For the state-updating digital twin, ordered anchors from each patient were grouped into sequences,

$$X^{\text{chunk}} \in \mathbb{R}^{L\times24\times4}. \tag{2}$$

Each anchor also included three structured modality groups: static variables representing demographic and clinical descriptors; biochemical variables representing laboratory measurements and biochemical markers; and contextual variables describing monitoring density, missingness, temporal gaps, elapsed-time information, freshness indicators, and recent glucose summary statistics. Structured variables were processed using median imputation followed by z-score standardization. We estimated all imputation values and normalization parameters exclusively from the training folds and applied them to the corresponding validation and test partitions.

### 3.4 Model Families

We compared three neural model families using the same anchor representation, preprocessing pipeline, and patient-level validation strategy: Hybrid LSTM–Transformer, Multimodal Digital Twin, and State-Updating Digital Twin. These architectures were selected to examine differences between target-specific sequence modeling, unified multimodal prediction, and longitudinal state updating. Conventional machine-learning models, including XGBoost and LightGBM, were explored during preliminary benchmarking and supplementary local attribution analyses but were not included in the primary comparative evaluation.

*Target-Specific Sequence Baseline.* The Hybrid LSTM–Transformer served as the principal sequence-based comparator. It combined bidirectional LSTM encoding with Transformer attention to capture temporal dependencies in the anchor-level input representation. Separate models were trained for each of the six prediction targets, using target-specific classification heads. This configuration provided a sequence-based reference for evaluating the two unified multitask digital-twin architectures.

*Multitask Multimodal Digital Twin.* The multimodal digital twin integrated temporal and structured inputs. The temporal branch used recurrent-transformer encoding, while static, biochemical, and contextual variables were processed through dedicated multilayer perceptrons. The resulting representations were fused into a shared latent vector,

$$z \in \mathbb{R}^{d_z}. \tag{3}$$

This shared representation was mapped to six task-specific outputs through a multitask prediction layer.

*State-Updating Digital Twin.* The state-updating digital twin enhanced the multimodal model by maintaining a longitudinal latent state across ordered anchors,

$$z_t = \mathrm{GRU}(e_t, z_{t-1}^{\mathrm{decay}}), \tag{4}$$

where $e_t$ represents the fused anchor-level embedding and $z_{t-1}^{\mathrm{decay}}$ denotes the gap-decayed previous latent state. Gap-dependent decay reduced the influence of older observations, while a validity mask prevented padded positions from affecting state updates or optimization.

### 3.5 Training and Patient-Level Validation

Model development and evaluation followed a five-fold patient-level cross-validation protocol. In each outer-fold experiment, one predefined patient fold was reserved for held-out testing, while the remaining four folds constituted the development dataset. The development anchors were globally ordered by their timestamps and divided into training (85%) and validation (15%) subsets using a chronological anchor-level split. This procedure ensured patient disjointness between the development and held-out test sets; however, because the internal split was performed at the anchor level rather than the patient level, individual patients could contribute observations to both training and validation subsets. The held-out test fold was excluded from model fitting, checkpoint selection, and classification-threshold optimization.

The three principal neural architectures were implemented in PyTorch and optimized using binary cross-entropy-based objectives with class-imbalance handling derived from the corresponding training data. The Hybrid LSTM–Transformer was trained as a target-specific binary prediction model, whereas the Multimodal Digital Twin and State-Updating Digital Twin were designed to generate six target-specific predictions within a multitask framework. Model-specific architectural configurations, optimization settings, and training hyperparameters are provided in Supplementary Table S2. Checkpoint selection was performed using validation performance without consulting the held-out test labels.

For each model, prediction target, and outer-fold experiment, the classification threshold was selected by maximizing the validation F1-score over a predefined grid ranging from 0.10 to 0.90 in increments of 0.05. The selected threshold was subsequently fixed and applied to predictions from the corresponding patient-disjoint held-out test fold. Threshold-dependent performance measures, including F1-score and balanced accuracy, were calculated using these validation-selected operating points, without re-optimizing thresholds on the held-out test labels. Threshold-independent discrimination measures, including AUROC and AUPRC, were computed directly from the predicted probabilities.

The resulting protocol provided patient-disjoint retrospective test evaluation while retaining a chronological, potentially patient-overlapping split for internal model selection. Consequently, the validation estimates should not be interpreted as independent patient-level generalization estimates. Similarly, the selected classification thresholds represent retrospectively optimized operating points for comparative evaluation rather than clinically validated alert thresholds. Prospective assessment and independent clinical validation would be required before deployment.

### 3.6 Predictive, Calibration, and Statistical Evaluation

Predictive performance was evaluated using AUROC, AUPRC, F1-score, and balanced accuracy. Because hypoglycemia was less frequent than hyperglycemia, we considered AUPRC and F1-score essential complements to AUROC. We assessed probability reliability using the Brier score, expected calibration error (ECE), and maximum calibration error (MCE) [10, 42, 43].

ECE and MCE were computed using 10 equal-width probability bins. For each bin, calibration error was calculated as the absolute difference between the mean predicted probability and the observed event frequency. ECE-10 was computed as the bin-size-weighted average calibration error, whereas MCE was defined as the maximum bin-level calibration error.

We compared the three principal neural model families pairwise using five paired outer-fold estimates. Wilcoxon signed-rank tests were performed for AUROC, AUPRC, F1-score, balanced accuracy, and Brier score for each prediction target. For each target–metric combination, the three pairwise model comparisons constituted a separate multiplicity family. Holm's step-down procedure was applied independently within each family to control the family-wise error rate at $\alpha = 0.05$. Given the five paired outer-fold estimates, the minimum attainable exact two-sided Wilcoxon signed-rank $p$-value was 0.0625 when all five paired differences were nonzero. Consequently, non-significant results were interpreted as insufficient evidence of predictive superiority rather than statistical equivalence.

### 3.7 Reliability, Explainability, and Behavioral Evaluation

The reliability evaluation was tailored to each model family. Sequence models were analyzed through temporal-position and channel attribution, the multimodal digital twin through modality-level attribution, and the state-updating digital twin through historical-anchor attribution.

We assessed explanation faithfulness by altering highly ranked explanatory components and measuring the resulting absolute change in predicted probability. Temporal perturbation involved modifying selected time positions or sequence channels, modality perturbation involved removing or replacing selected modality branches, and state-history perturbation involved removing or masking selected historical anchors. Faithfulness was summarized as the mean absolute probability change across targets. Stability was evaluated by assessing the consistency of the dominant explanatory component across folds and targets under each model-specific explanation protocol.

We quantified cross-model disagreement by examining differences in binary decisions and probability outputs among the three neural models. Errors were descriptively stratified by demographic category, biochemical-record availability, monitoring density, missingness level, and temporal-gap characteristics. We computed error rates within each stratum and compared them across model families. These exploratory analyses were not interpreted as a formal fairness assessment. Behavioral robustness was assessed through controlled perturbations of temporal history, structured modalities, and state sequences. Recent-glucose perturbation involved lowering recent glucose values to assess sensitivity to hypoglycemia and increasing recent glucose values to assess sensitivity to hyperglycemia. Structured-modality perturbation involved replacing biochemical or contextual branches with neutral or training-distribution values. State-history perturbation removed or masked selected historical anchors in the state-updating model. Changes in predicted probability were used as behavioral sensitivity measures, not as causal or clinically actionable counterfactuals.

### 3.8 Reproducibility and Software Considerations

The framework was implemented using Python, PyTorch, NumPy, pandas, and scikit-learn. Preprocessing, model training, checkpoint selection, threshold optimization, and evaluation were performed independently within each patient-level fold to prevent information leakage. The T1DiabetesGranada dataset is publicly available [34]. Supplementary Table S2 summarizes the principal model architectures, loss function, optimizer choices, class-weighting strategy, checkpoint selection, and training protocol. Full computational reproducibility additionally requires the exact feature definitions, patient-level fold assignments, preprocessing settings, numerical hyperparameters, and evaluation code, which are not completely specified in the current reporting.

Computational efficiency was not included as a model-selection criterion because training time, memory consumption, and inference latency were not measured consistently across all experimental stages. Consequently, the study does not establish comparative computational efficiency between independently trained sequence models and unified multitask digital twins.

## 4 Analysis of the Results

### 4.1 RQ1: Anchor-Based Multimodal Risk Representation

RQ1 examined whether heterogeneous longitudinal glucose, static, biochemical, diagnostic, missingness, and temporal-gap information could be transformed into a unified anchor-based representation for joint short-term hypoglycemia and hyperglycemia prediction. The resulting analytical dataset retained the full cohort and preserved variation in monitoring density and modality availability. It established common prediction units, retrospective windows, outcome definitions, and patient-level validation boundaries for all model families.

*Cohort Coverage and Multimodal Availability.* The cleaned cohort comprised 736 individuals with type 1 diabetes, all with longitudinal glucose records. Biochemical information was available for 723 patients (98.23%), diagnostic information for 511 patients (69.43%), and both modalities for 498 patients. No patient had glucose-only data. Following cleaning and harmonization, the dataset contained 22,649,679 glucose measurements, 87,470 biochemical records, and 1,733 diagnostic records. Cohort coverage is summarized in Table 1.

Table 1. Cohort coverage after data cleaning and harmonization.

| **Cohort characteristic** | **Value** |
|---|---|
| Total patients | 736 |
| Patients with glucose records | 736 |
| Patients with biochemical records | 723 |
| Patients with diagnostic records | 511 |
| Patients with both biochemical and diagnostic records | 498 |
| Patients with glucose-only records | 0 |
| Cleaned glucose measurements | 22,649,679 |
| Cleaned biochemical records | 87,470 |
| Cleaned diagnostic records | 1,733 |
| Biochemical coverage | 98.23% |
| Diagnostic coverage | 69.43% |
| Mean glucose measurements per patient | 30,774.02 |

Longitudinal monitoring volume varied substantially across patients. The median number of glucose measurements was 24,993.5, with an interquartile range of 10,668.75–45,684.75 and an overall range of 401–137,275. This variation motivated us to retain monitoring-density, missingness, and temporal-gap descriptors in the common representation. The corresponding modality-coverage visualization is provided in Supplementary Fig. S1.

*Anchor-Level Dataset Construction.* The harmonized records produced 5,503,441 prediction anchors covering all 736 patients. Each anchor represented a reference prediction time and included a six-hour retrospective glucose history sampled at 15-minute resolution, yielding 24 temporal positions. We generated candidate anchors at a stride of four temporal steps, corresponding to a nominal one-hour interval. Anchors were retained when at least five glucose measurements were available within the retrospective window, and a glucose value was observed at the anchor timestamp. This criterion excluded anchors with insufficient recent evidence while preserving complete cohort coverage. Because future-risk outcomes were defined from observed glucose measurements, interpreting each label also depended on the availability of future observations within its corresponding prediction horizon. The final analytical representation included 43 variables: identifiers, temporal summaries, monitoring-density measures, missingness and gap descriptors, static attributes, biochemical variables, diagnostic summaries, and six future-risk outcomes. The same anchor construction was applied to all model families. Table 2 summarizes the resulting analytical dataset.

The five outer folds contained 147–148 patients and approximately 1.03–1.18 million held-out anchors each. Differences in anchor volume reflected variation in monitoring duration, data density, and anchor eligibility. Assigning all anchors from the same patient to one fold prevented overlapping retrospective histories and future observation intervals from appearing across development and held-out test sets.

Although the analytical dataset contained more than 5.5 million anchors, these represented repeated and partially overlapping longitudinal observations rather than independent samples. The independent patient cohort therefore remained 736 individuals. The detailed fold-level anchor distribution is shown in Supplementary Fig. S2.

Table 2. Summary of the anchor-level multimodal analytical dataset.

| **Dataset property** | **Value** |
|---|---|
| Anchor-level samples | 5,503,441 |
| Patients represented | 736 |
| Final analytical variables | 43 |
| Retrospective observation window | 6 hours |
| Temporal resolution | 15 minutes |
| Temporal positions per anchor | 24 |
| Temporal channels per position | 4 |
| Minimum observed glucose values per window | 5 |
| Anchor stride | 4 temporal steps |
| Prediction horizons | 30, 60, and 120 minutes |
| Binary prediction targets | 6 |
| Outer validation folds | 5 patient-level folds |

*Temporal Irregularity and Intermeasurement Gaps.* The longitudinal glucose records contained 22,648,943 consecutive intermeasurement gaps. The first quartile, median, and third quartile were all 15 minutes, indicating that most observations followed the nominal CGM sampling interval. However, the mean gap was 20.76 minutes, and the maximum reached 1,516,981 minutes, revealing a strongly skewed distribution with rare prolonged interruptions.

Figure 2 illustrates this contrast. The observed irregularity supports including observation masks, elapsed-time variables, freshness indicators, inter-anchor gaps, and gap-dependent state decay.

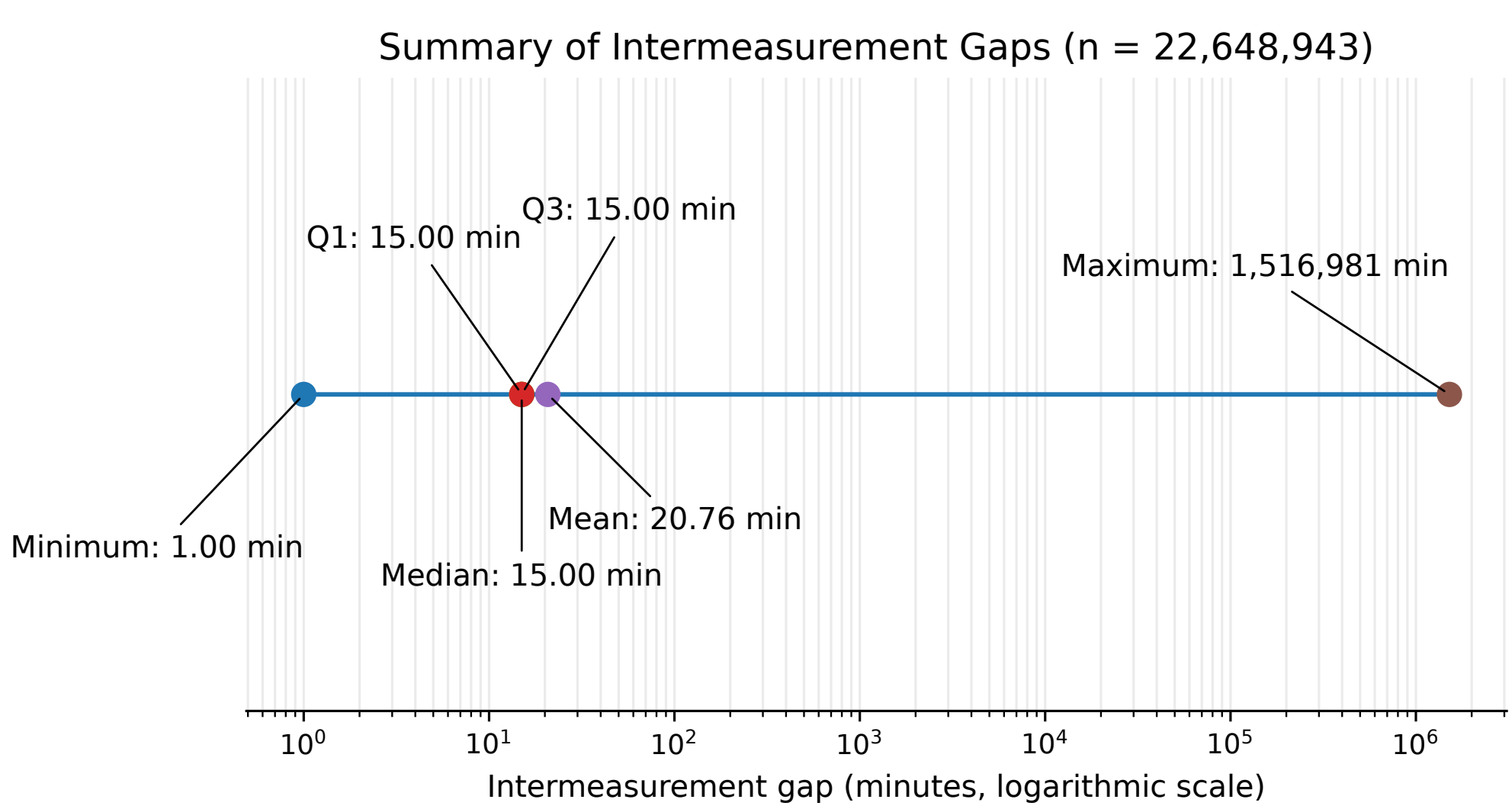

Fig. 2. Glucose intermeasurement gaps on a logarithmic scale. Most observations followed the nominal 15-minute interval, whereas rare prolonged interruptions produced a strongly skewed upper tail.

Supplementary Table S3 reports complete descriptive statistics for intermeasurement gaps.

*Outcome Distribution Across Prediction Horizons.* Each anchor was assigned six binary outcomes corresponding to hypoglycemia and hyperglycemia within 30-, 60-, and 120-minute horizons. Hypoglycemia was defined as at least one observed future glucose value below 70 mg/dL, whereas hyperglycemia was defined as at least one observed future value above 180 mg/dL within the corresponding horizon. These outcomes represent observed glycemic events and depend on the availability of future glucose measurements.

Table 3 reports the counts and rates of positive outcomes. Hypoglycemia prevalence increased from 5.76% at 30 minutes to 11.07% at 120 minutes, while hyperglycemia prevalence increased from 37.98% to 50.23%.

Table 3. Distribution of positive glycemic-risk outcomes across prediction horizons.

| **Prediction target** | **Positive anchors** | **Positive rate (%)** |
|---|---:|---:|
| Hypoglycemia within 30 min | 316,853 | 5.76 |
| Hypoglycemia within 60 min | 421,630 | 7.66 |
| Hypoglycemia within 120 min | 608,956 | 11.07 |
| Hyperglycemia within 30 min | 2,090,020 | 37.98 |
| Hyperglycemia within 60 min | 2,336,472 | 42.45 |
| Hyperglycemia within 120 min | 2,764,473 | 50.23 |

Hypoglycemia remained substantially more imbalanced than hyperglycemia at every horizon, while the 120-minute hyperglycemia target approached a balanced distribution. This class structure supported using AUPRC, F1-score, balanced accuracy, calibration measures, and threshold-dependent evaluation in addition to AUROC. A visual comparison of target prevalence is provided in Supplementary Fig. S3.

| **Answer to RQ1** | The heterogeneous T1DiabetesGranada records were transformed into 5,503,441 standardized prediction anchors from 736 patients, supporting hypoglycemia and hyperglycemia prediction at 30-, 60-, and 120-minute horizons. The unified representation preserved multimodal availability, temporal irregularity, monitoring heterogeneity, and target imbalance while maintaining patient-level validation boundaries. Because anchors represented repeated longitudinal observations, outcome interpretation remained dependent on future glucose observation availability. |
|---|---|

## 4.2 RQ2: Comparative Predictive Performance

RQ2 evaluated whether multitask multimodal fusion and longitudinal state updating provided measurable predictive advantages over the target-specific Hybrid LSTM–Transformer under the same anchor construction, six prediction targets, preprocessing pipeline, patient-level folds, and held-out evaluation sets.

*Overall and Horizon-Dependent Performance.* Table 4 reports the mean and standard deviation of AUROC, AUPRC, and F1-score across the five patient-level outer folds. All three model families achieved their strongest performance for the 30-minute outcomes, followed by progressive reductions at 60 and 120 minutes. Differences among the architectures were generally small relative to their fold-level variability.

The digital-twin models achieved several small numerical advantages at 30 and 60 minutes. For 30-minute hypoglycemia, the Multimodal Digital Twin and State-Updating Digital Twin achieved identical mean AUROC values at the reported precision. At the same time, the Multimodal Digital Twin obtained a marginally higher mean AUPRC. The Multimodal Digital Twin also achieved the highest mean AUROC, AUPRC, and F1-score for 30-minute hyperglycemia. The State-Updating Digital Twin achieved the highest mean AUPRC and F1-score for 60-minute hypoglycemia. In contrast, the Hybrid LSTM–Transformer achieved the highest mean AUROC, AUPRC, and F1-score for both 120-minute outcomes.

Table 4. Predictive performance of the principal neural model families across the six glycemic-risk targets. Results are reported as mean ± standard deviation across five patient-level outer folds. HLT: Hybrid LSTM–Transformer; MMDT: Multimodal Digital Twin; SUDT: State-Updating Digital Twin; Hypo: hypoglycemia; Hyper: hyperglycemia.

| Target | HLT | | | MMDT | | | SUDT | | |
|---|---|---|---|---|---|---|---|---|---|
| | **AUROC** | **AUPRC** | **F1** | **AUROC** | **AUPRC** | **F1** | **AUROC** | **AUPRC** | **F1** |
| Hypo 30 min | 0.9877 ±0.0013 | 0.8621 ±0.0111 | 0.7824 ±0.0096 | 0.9884 ±0.0010 | 0.8692 ±0.0103 | 0.7874 ±0.0107 | 0.9884 ±0.0009 | 0.8691 ±0.0117 | 0.7876 ±0.0115 |
| Hypo 60 min | 0.9624 ±0.0029 | 0.7828 ±0.0183 | 0.7013 ±0.0156 | 0.9630 ±0.0025 | 0.7850 ±0.0158 | 0.7012 ±0.0147 | 0.9629 ±0.0029 | 0.7857 ±0.0182 | 0.7052 ±0.0155 |
| Hypo 120 min | 0.9017 ±0.0069 | 0.6809 ±0.0281 | 0.6125 ±0.0202 | 0.8986 ±0.0056 | 0.6774 ±0.0223 | 0.6035 ±0.0191 | 0.8985 ±0.0090 | 0.6789 ±0.0261 | 0.6078 ±0.0196 |
| Hyper 30 min | 0.9880 ±0.0005 | 0.9838 ±0.0013 | 0.9315 ±0.0028 | 0.9883 ±0.0004 | 0.9843 ±0.0012 | 0.9325 ±0.0026 | 0.9882 ±0.0004 | 0.9841 ±0.0012 | 0.9323 ±0.0026 |
| Hyper 60 min | 0.9613 ±0.0010 | 0.9619 ±0.0028 | 0.8912 ±0.0040 | 0.9606 ±0.0013 | 0.9619 ±0.0030 | 0.8916 ±0.0043 | 0.9610 ±0.0012 | 0.9620 ±0.0029 | 0.8918 ±0.0043 |
| Hyper 120 min | 0.9087 ±0.0016 | 0.9319 ±0.0049 | 0.8340 ±0.0058 | 0.9049 ±0.0024 | 0.9302 ±0.0053 | 0.8320 ±0.0065 | 0.9062 ±0.0022 | 0.9308 ±0.0054 | 0.8330 ±0.0060 |

The magnitude of these differences was limited. The spread between the highest and lowest mean F1-scores was approximately 0.005 for 30-minute hypoglycemia, 0.004 for 60-minute hypoglycemia, and less than 0.001 for 60-minute hyperglycemia. Detailed AUPRC and F1 comparisons are provided in Supplementary Figs. S4 and S5.

A consistent horizon-dependent decline was observed across all model families. For hypoglycemia, mean F1-score decreased from approximately 0.78–0.79 at 30 minutes to 0.70–0.71 at 60 minutes and 0.60–0.61 at 120 minutes. Hyperglycemia F1-score decreased from approximately 0.932 at 30 minutes to approximately 0.892 at 60 minutes and 0.832–0.834 at 120 minutes. Figure 3 shows nearly parallel trajectories across models. The 120-minute hypoglycemia outcome was the most difficult task, as indicated by the reported AUPRC and F1-score values across all three architectures.

*Target-Specific Rankings and Statistical Comparison.* Table 5 summarizes the model with the highest numerical mean for each target and metric. No architecture achieved the highest mean across all outcomes.

The Multimodal Digital Twin achieved the highest mean values for the 30-minute outcomes, whereas the State-Updating Digital Twin achieved the highest mean AUPRC and F1-score for both 60-minute outcomes. The Hybrid LSTM–Transformer was numerically strongest for both 120-minute outcomes. A visual winner map is provided in Supplementary Fig. S6.

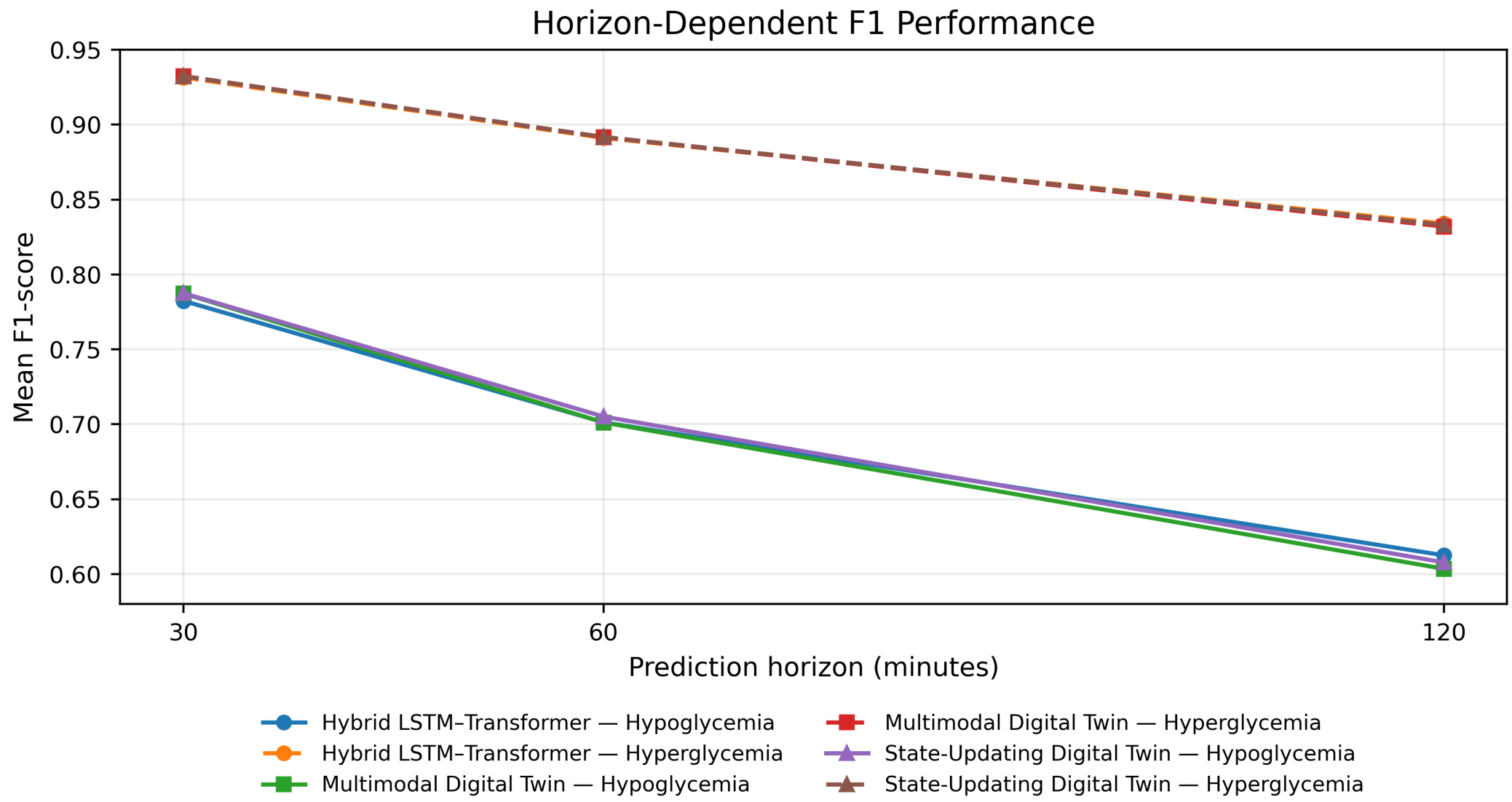

Fig. 3. Horizon-dependent mean F1-score for hypoglycemia and hyperglycemia prediction. Performance declined from 30 to 120 minutes for all model families, with a larger reduction for hypoglycemia.

Table 5. Target-wise model ranking based on mean predictive performance. Rankings indicate numerical differences and do not imply statistically established superiority. Ties are reported at the displayed numerical precision. HLT: Hybrid LSTM–Transformer; MMDT: Multimodal Digital Twin; SUDT: State-Updating Digital Twin.

| **Target** | **Best mean AUROC** | **Best mean AUPRC** | **Best mean F1-score** |
|---|---|---|---|
| Hypoglycemia 30 min | MMDT / SUDT | MMDT | SUDT |
| Hypoglycemia 60 min | MMDT | SUDT | SUDT |
| Hypoglycemia 120 min | HLT | HLT | HLT |
| Hyperglycemia 30 min | MMDT | MMDT | MMDT |
| Hyperglycemia 60 min | HLT | SUDT | SUDT |
| Hyperglycemia 120 min | HLT | HLT | HLT |

Pairwise fold-level comparisons were conducted using Wilcoxon signed-rank tests with Holm correction across the evaluated model–metric combinations. No model family achieved a Holm-corrected statistically significant advantage for AUROC, AUPRC, F1-score, balanced accuracy, or Brier score on any of the six targets. However, statistical inference was constrained by the availability of only five paired outer-fold estimates. For five nonzero paired differences, the minimum attainable two-sided exact Wilcoxon signed-rank p-value is 0.0625, which exceeds the conventional significance threshold of $\alpha = 0.05$ even before multiple-comparison correction. Consequently, the lack of statistically significant differences reflects the test's limited resolution. It should not be interpreted as evidence of statistical equivalence or established similarity between the models.

Overall, the two digital-twin formulations achieved numerically similar predictive performance to the Hybrid LSTM–Transformer while producing all six risk estimates within a unified multitask framework. Increasing the prediction horizon was associated with a larger descriptive reduction in predictive performance than the differences observed among model architectures.

**Answer to RQ2**

Under the common anchor-based, patient-level evaluation protocol, the digital-twin models achieved small numerical gains at 30- and 60-minute horizons, while the Hybrid LSTM–Transformer performed best for both 120-minute outcomes. However, no statistically significant predictive advantage was established, with inference limited by five paired outer-fold estimates. These findings support multimodal and state-updating digital twins as competitive unified multitask alternatives without establishing predictive superiority or equivalence.

### 4.3 RQ3: Reliability, Explainability, and Deployment-Oriented Behavior

RQ3 investigated how the three principal model families differed beyond predictive discrimination, with particular emphasis on probabilistic reliability, threshold-dependent classification performance, explanation faithfulness and stability, cross-model disagreement, subgroup error patterns, and responses to controlled input perturbations. Given the limited statistical resolution of the RQ2 comparisons, these analyses were intended to characterize complementary model properties rather than establish architectural superiority or clinical deployment readiness.

*Calibration and Threshold-Dependent Performance.* Table 6 summarizes the mean Brier score and expected calibration error (ECE-10) across the six prediction targets. The Multimodal Digital Twin achieved the lowest mean Brier score (0.0702), whereas the State-Updating Digital Twin achieved the lowest mean ECE-10 (0.0611). The Brier score quantifies overall probabilistic prediction error, incorporating both calibration and discrimination, whereas ECE-10 estimates the discrepancy between predicted probabilities and observed event frequencies under a 10-bin calibration scheme. These metrics therefore characterize related but distinct aspects of probabilistic reliability.

Table 6. Mean probabilistic reliability metrics across the six prediction targets. Lower values indicate better numerical performance under the corresponding metric. Bold values identify the lowest mean for each metric and do not imply statistically significant superiority.

| **Model family** | **Mean Brier** | **Mean ECE-10** |
|---|---|---|
| Hybrid LSTM–Transformer | 0.0763 | 0.0741 |
| Multimodal Digital Twin | **0.0702** | 0.0658 |
| State-Updating Digital Twin | 0.0709 | **0.0611** |

Calibration discrepancies were most pronounced for longer-horizon hypoglycemia prediction. At 120 minutes, ECE-10 reached 0.1845 for the Hybrid LSTM–Transformer, 0.1715 for the Multimodal Digital Twin, and 0.1633 for the State-Updating Digital Twin. In contrast, ECE-10 remained below 0.026 across all hyperglycemia targets and model families. Both digital-twin formulations achieved lower mean ECE-10 values than the sequence baseline, while the Multimodal Digital Twin achieved the lowest mean Brier score. These descriptive differences indicate distinct probabilistic reliability profiles but do not establish statistically significant or clinically meaningful calibration superiority.

Table 7. Balanced accuracy of the three principal neural model families across six glycemic-risk prediction targets. Values are reported as mean ± standard deviation across five patient-level outer folds. Task-specific classification thresholds were selected by maximizing validation F1-score within each outer-fold experiment and fixed before evaluation on the corresponding patient-disjoint held-out test set. Bold values indicate the highest numerical mean for each target and do not imply statistically significant superiority.

| Prediction target | Hybrid LSTM–Transformer | Multimodal Digital Twin | State-Updating Digital Twin |
|---|---|---|---|
| Hypoglycemia 30 min | **0.8887 ± 0.0117** | 0.8863 ± 0.0056 | 0.8857 ± 0.0086 |
| Hypoglycemia 60 min | 0.8194 ± 0.0111 | 0.8212 ± 0.0146 | **0.8285 ± 0.0085** |
| Hypoglycemia 120 min | **0.7699 ± 0.0116** | 0.7538 ± 0.0107 | 0.7604 ± 0.0105 |
| Hyperglycemia 30 min | 0.9434 ± 0.0011 | **0.9440 ± 0.0013** | 0.9439 ± 0.0012 |
| Hyperglycemia 60 min | 0.9042 ± 0.0015 | 0.9046 ± 0.0020 | **0.9047 ± 0.0021** |
| Hyperglycemia 120 min | **0.8405 ± 0.0015** | 0.8378 ± 0.0042 | 0.8387 ± 0.0034 |

Threshold-dependent classification performance was evaluated using task-specific decision thresholds selected independently within each outer-fold experiment. Thresholds were chosen by maximizing F1-score on validation predictions and subsequently fixed before evaluation on the corresponding patient-disjoint held-out test sets. Balanced accuracy was calculated using these validation-selected thresholds, although it was not itself the threshold-selection objective.

Table 7 reports the resulting balanced accuracy as mean ± standard deviation across the five outer folds. Unlike a retrospective threshold sweep performed directly on pooled evaluation predictions, this analysis preserves the separation between threshold selection and held-out performance assessment. The reported values therefore characterize threshold-dependent classification performance under the study's retrospective evaluation protocol rather than clinically validated alert operating points.

For hypoglycemia prediction, the Hybrid LSTM–Transformer achieved the highest mean balanced accuracy at 30 minutes (0.8887) and 120 minutes (0.7699), whereas the State-Updating Digital Twin achieved the highest value at 60 minutes (0.8285). For hyperglycemia prediction, the Multimodal Digital Twin achieved the highest mean balanced accuracy at 30 minutes (0.9440), the State-Updating Digital Twin at 60 minutes (0.9047), and the Hybrid LSTM–Transformer at 120 minutes (0.8405).

Across all three architectures, balanced accuracy decreased as the prediction horizon increased. Differences among the model families were relatively small for hyperglycemia, particularly at 30 and 60 minutes. These findings indicate that threshold-dependent numerical rankings varied across prediction targets and horizons, without establishing statistically supported architectural superiority. Furthermore, because decision thresholds were selected to maximize validation F1-score rather than clinical utility, the resulting operating points should not be interpreted as clinically optimized alert settings. Prospective evaluation would be necessary to determine appropriate thresholds based on sensitivity, specificity, false-alert burden, and the consequences of missed glycemic events.

*Global and Model-Specific Explanation Structure.* Explanation analysis was adapted to the internal representation of each model family. For the Hybrid LSTM–Transformer, attribution was evaluated over temporal positions and sequence channels. For the Multimodal Digital Twin, attribution was evaluated over temporal, static, biochemical, and contextual branches. For the State-Updating

Digital Twin, attribution was evaluated over ordered historical anchors. These explanation units reflect the respective model architectures and should not be interpreted as directly interchangeable.

Figure 4 shows that Hybrid LSTM–Transformer attribution was concentrated near the most recent positions of the six-hour retrospective window. Some longer-horizon outcomes distributed importance over broader temporal intervals, although recent glucose observations remained dominant.

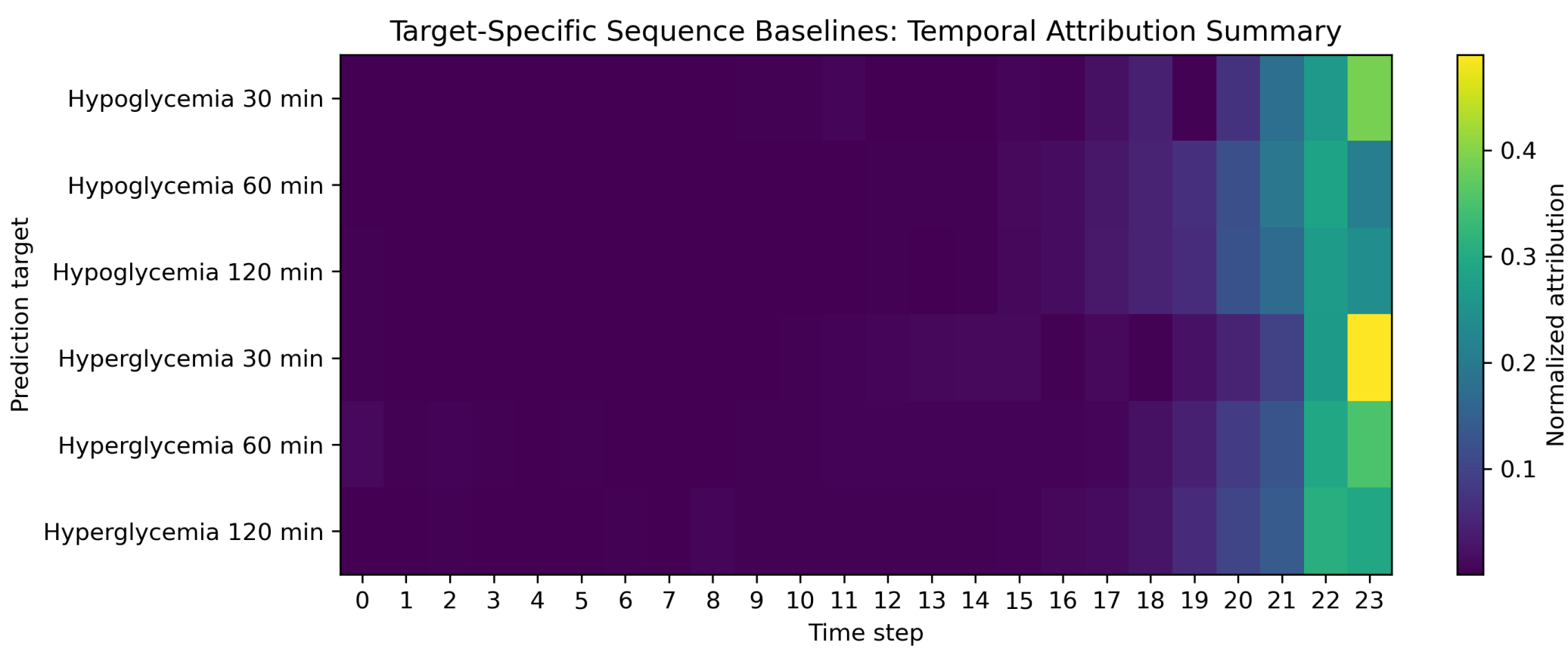

Fig. 4. Normalized temporal attribution for the Hybrid LSTM–Transformer across the six prediction targets. Importance was concentrated near the most recent retrospective positions, while some longer-horizon outcomes showed broader temporal distributions.

Observation-mask, elapsed-time, and freshness channels contributed smaller but non-zero amounts, indicating that measurement availability and recency also influenced sequence-model predictions.

The Multimodal Digital Twin exhibited target-dependent modality contributions. As shown in Fig. 5, the temporal branch dominated most hyperglycemia outcomes and imminent hypoglycemia, whereas contextual information contributed more strongly to some longer-horizon hypoglycemia targets. Static and biochemical branches generally showed smaller individual-anchor contributions.

For the State-Updating Digital Twin, attribution was distributed across multiple historical anchors. Recent and target-proximal anchors frequently received high importance, although the dominant anchor varied across folds and prediction targets. This pattern suggests that the model's recurrent state representation may distribute predictive information across multiple historical observations rather than relying exclusively on the most recent anchor.

Table 8 summarizes the dominant global explanation patterns across the three architectures.

Supplementary Section A.5.1 presents representative true-positive, false-positive, false-negative, and true-negative local SHAP explanations for selected hyperglycemia prediction cases.

*Explanation Faithfulness and Stability.* Explanation faithfulness was assessed by perturbing highly ranked explanatory components and measuring the resulting absolute change in predicted probability. Explanation stability characterized the consistency of the dominant explanatory component across folds and prediction targets.

Because the evaluated components differed across temporal positions, modality branches, and historical anchors, the resulting values reflect model-specific explanation protocols rather than a shared measurement scale. Consequently, they should not be used to establish a direct ranking of explanation quality across architectures.

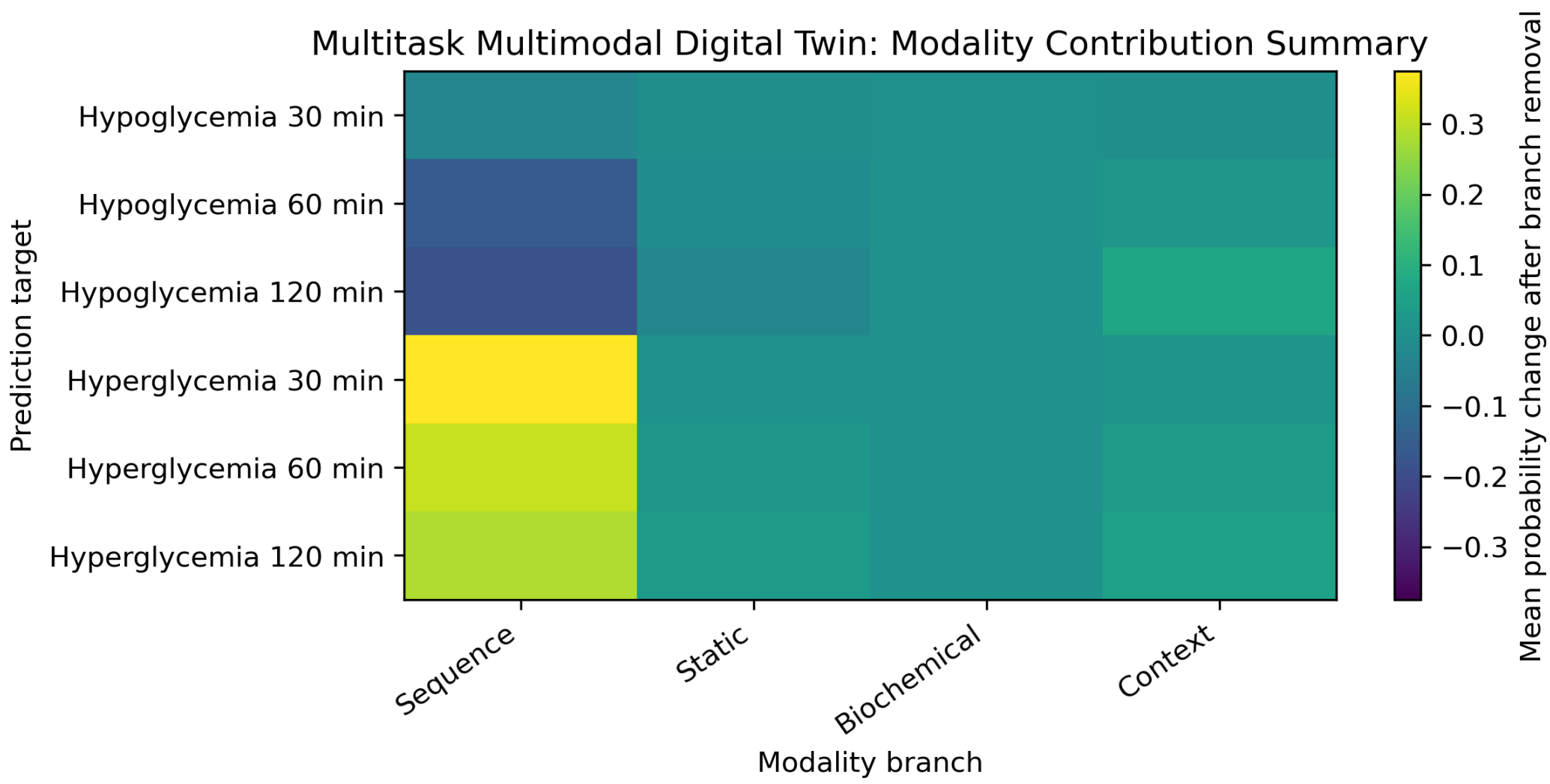

Fig. 5. Modality-level contribution in the Multimodal Digital Twin. The temporal branch dominated most targets, while contextual information contributed more strongly to some longer-horizon hypoglycemia outcomes.

Table 8. Dominant global explanation patterns across the principal model families.

| **Model family** | **Explanation unit** | **Dominant global pattern** | **Target-dependent behavior** |
|---|---|---|---|
| Hybrid LSTM–Transformer | Temporal positions and sequence channels | Recent glucose values and final retrospective positions | Longer-horizon targets distributed importance over a broader temporal interval. |
| Multimodal Digital Twin | Temporal, static, biochemical, and contextual branches | Temporal sequence branch | Contextual contribution increased for some longer-horizon hypoglycemia targets. |
| State-Updating Digital Twin | Ordered historical anchors | Recent and target-proximal anchors with distributed importance | Dominant anchor positions varied across folds and targets. |

Table 9 summarizes the mean perturbation effect and dominant-component stability under the respective protocols.

Under its temporal explanation protocol, the Hybrid LSTM–Transformer produced a mean perturbation effect of 0.3105 and dominant-component stability of 1.0000. The Multimodal Digital Twin showed a mean modality-level perturbation effect of 0.1722 and stability of 0.9667. The State-Updating Digital Twin exhibited a mean individual-anchor perturbation effect of 0.0074 and dominant-anchor stability of 0.3667.

These findings describe distinct attribution structures rather than establishing a common ranking of explanation quality. In particular, the smaller individual-anchor perturbation effects and lower dominant-anchor consistency observed for the State-Updating Digital Twin may reflect information distributed across recurrent state updates. However, this interpretation remains descriptive and does not demonstrate that the model is intrinsically less reliable or that its explanations are less faithful. Target-specific faithfulness and stability results are presented in Supplementary Fig. S8.

Table 9. Summary of model-specific explanation faithfulness and stability. Faithfulness denotes the mean absolute selected target-level perturbation effect across the six prediction targets. Stability denotes dominant-component consistency under the model-specific definition. Values are descriptive and are not directly comparable across different explanation units.

| Model family | Faithfulness effect | Stability |
|---|---|---|
| Hybrid LSTM–Transformer | 0.3105 | 1.0000 |
| Multimodal Digital Twin | 0.1722 | 0.9667 |
| State-Updating Digital Twin | 0.0074 | 0.3667 |

*Cross-Model Disagreement and Error Patterns.* Cross-model disagreement was examined to characterize differences in model outputs beyond conventional predictive metrics. Table 10 reports binary disagreement rates and mean probability spread across the six prediction targets. These measures quantify variation among model predictions rather than prediction errors or calibrated uncertainty.

Table 10. Cross-model disagreement and mean probability spread across prediction targets.

| Target | Disagreement rate (%) | Mean probability spread |
|---|---|---|
| Hypoglycemia 30 min | 1.27 | 0.0402 |
| Hypoglycemia 60 min | 1.95 | 0.0699 |
| Hypoglycemia 120 min | 3.89 | 0.1080 |
| Hyperglycemia 30 min | 1.74 | 0.0343 |
| Hyperglycemia 60 min | 2.82 | 0.0494 |
| Hyperglycemia 120 min | 6.60 | 0.0657 |

Binary disagreement increased with prediction horizon for both hypoglycemia and hyperglycemia. The highest disagreement rate occurred for 120-minute hyperglycemia (6.60%), whereas the largest mean probability spread was observed for 120-minute hypoglycemia (0.1080).

The unweighted mean disagreement across the three horizons was 2.37% for hypoglycemia and 3.72% for hyperglycemia. Thus, the architectures agreed on most binary decisions, although divergence increased at longer prediction horizons. These observations should be interpreted in relation to the thresholding procedure and disagreement definitions used in the evaluation protocol.

Descriptive subgroup analyses showed no consistent elevation in error rates across age groups. Slightly higher error rates were observed among patients without biochemical records, although these differences may reflect variation in monitoring history, clinical characteristics, or overall data completeness. Supplementary Fig. S10 provides detailed subgroup results. These analyses were exploratory and should not be interpreted as a formal assessment of algorithmic fairness.

*Controlled Perturbation and State-History Behavior.* Controlled input perturbations were used to examine model sensitivity to modifications in recent glucose, observation missingness, structured variables, and historical anchors. These experiments characterized prediction behavior under predefined input changes rather than causal responses or physiological effects.

Figure 6 summarizes the responses to perturbations in recent glucose.

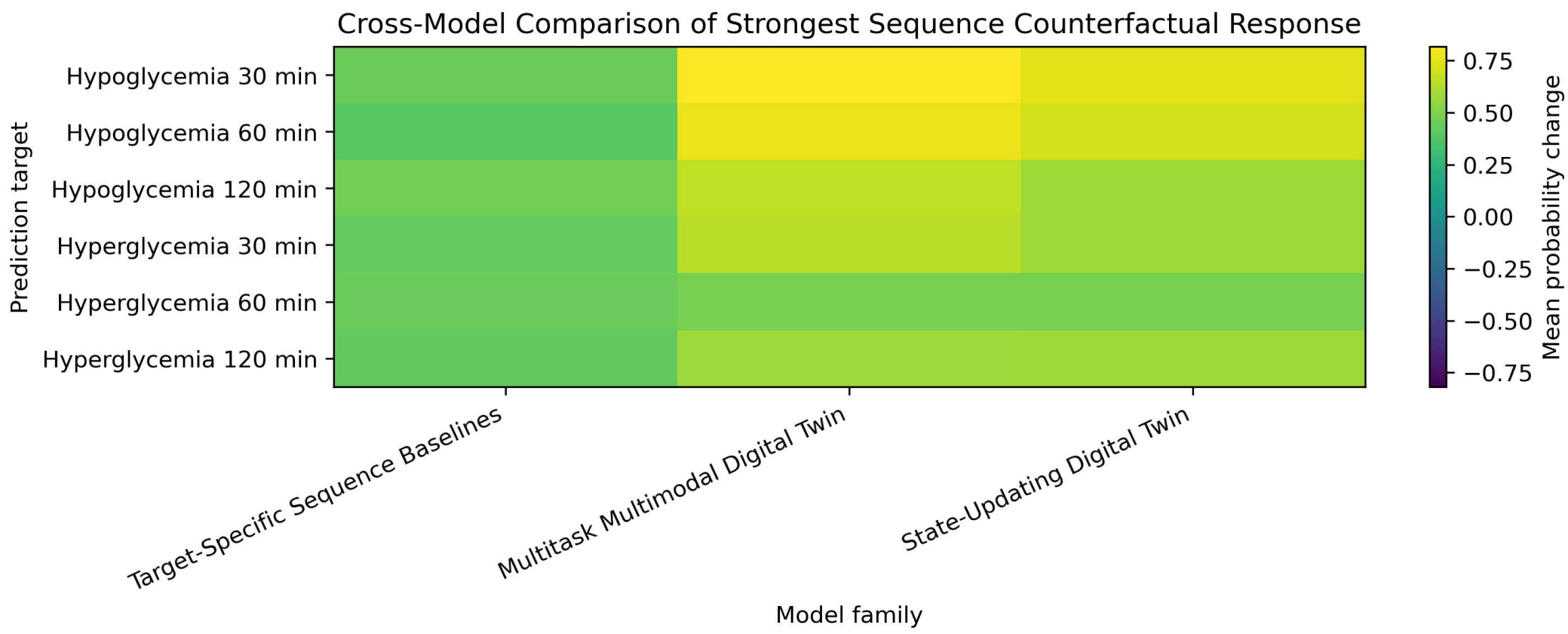

Fig. 6. Recent-sequence perturbation responses across model families. Lowering recent glucose increased predicted hypoglycemia risk, whereas increasing recent glucose increased predicted hyperglycemia risk. These changes represent behavioral sensitivity rather than causal effects or demonstrated robustness.

Across all architectures, lowering recent glucose increased predicted hypoglycemia risk, whereas increasing recent glucose increased predicted hyperglycemia risk. For the Hybrid LSTM–Transformer, mean absolute probability changes ranged from 0.401 to 0.466 for hypoglycemia and from 0.417 to 0.445 for hyperglycemia. The corresponding ranges were 0.661–0.818 and 0.484–0.641 for the Multimodal Digital Twin, and 0.583–0.751 and 0.480–0.586 for the State-Updating Digital Twin.

These responses were substantially larger than those associated with individual biochemical or contextual perturbations, which generally produced probability changes below 0.012. Detailed results are provided in Supplementary Table S4. The magnitude and direction of these responses indicate sensitivity to the modified inputs but do not independently establish model robustness, causal validity, or clinical appropriateness.

Increasing recent missingness also modified model outputs, indicating sensitivity to observation masks, elapsed-time information, and freshness channels. Such responses may reflect adaptation to reduced information availability, dependence on dataset-specific recording patterns, or a combination of both.

For the State-Updating Digital Twin, removing individual historical anchors produced smaller and non-monotonic changes across prediction horizons. This behavior is consistent with predictive information being distributed across recurrent state updates rather than concentrated in a single historical anchor. Nevertheless, alternative explanations cannot be excluded, and the findings do not establish the physiological validity of the learned state representation. Detailed state-history stress-test results are provided in Supplementary Fig. S11.

*Deployment-Oriented Synthesis.* Table 11 integrates probabilistic reliability, joint prediction capability, longitudinal state updating, model-specific explanation stability, and principal architectural limitations. The synthesis identifies characteristics relevant to future deployment-oriented investigation rather than demonstrating clinical implementation readiness.

The Hybrid LSTM–Transformer demonstrated substantial temporal-component perturbation effects and high dominant-component consistency under its model-specific explanation protocol but required separate models for the six prediction targets. The Multimodal Digital Twin combined joint six-task prediction with the lowest mean Brier score and high modality-level dominant-component consistency. The State-Updating Digital Twin achieved the lowest mean ECE-10 and supported joint prediction through explicit longitudinal state propagation, gap-dependent updating, and validity masking.

Table 11. Integrated deployment-oriented comparison of the principal model families. Explanation stability values are model-specific and should not be directly ranked across architectures. Bold values indicate the lowest numerical mean for each calibration metric and do not imply statistically significant superiority. HLT: Hybrid LSTM–Transformer; MMDT: Multimodal Digital Twin; SUDT: State-Updating Digital Twin.

| Model | Mean Brier | Mean ECE-10 | Joint six-task output | Longitudinal state update | Stability | Primary limitation |
|---|---|---|---|---|---|---|
| HLT | 0.0763 | 0.0741 | No | No | 1.0000 | Requires a separate model for each prediction target. |
| MMDT | **0.0702** | 0.0658 | Yes | No | 0.9667 | Does not explicitly propagate latent states across consecutive prediction anchors. |
| SUDT | 0.0709 | **0.0611** | Yes | Yes | 0.3667 | Lower dominant-component attribution consistency and mixed responses to history-stress tests. |

Because explanation units and perturbation protocols differed among architectures, their stability and faithfulness values do not establish a common interpretability ranking. Similarly, numerical differences in calibration and balanced accuracy should not be interpreted as evidence of statistically established superiority.

The State-Updating Digital Twin was identified as a candidate for further deployment-oriented investigation because it combined joint six-task inference, explicit longitudinal updating, and the lowest mean ECE-10. However, these characteristics do not establish overall predictive superiority, greater clinical utility, or deployment readiness. External validation, prospective assessment, clinical threshold selection, and evaluation of computational and operational requirements remain necessary before considering practical deployment.

**Answer to RQ3**

The three model families exhibited distinct reliability and architectural profiles despite broadly similar predictive performance. The Multimodal Digital Twin achieved the lowest mean Brier score, whereas the State-Updating Digital Twin achieved the lowest mean ECE-10 and supported explicit longitudinal state updating. Balanced accuracy declined with increasing prediction horizon, while numerical model rankings varied across targets. Explanation analyses revealed architecture-specific attribution patterns that were not directly comparable across model families. Cross-model disagreement increased with prediction horizon, and controlled glucose perturbations produced directionally consistent prediction responses. Collectively, these findings identify complementary model characteristics and areas requiring further validation rather than establishing architectural superiority or clinical deployment readiness.

## 5 Discussion

This study introduced an anchor-based comparative framework for short-term glycemic-risk prediction that uses longitudinal glucose trajectories and multimodal patient information. By standardizing prediction anchors, outcome definitions, preprocessing, and patient-level validation, the framework facilitated a controlled comparison of a target-specific Hybrid LSTM–Transformer, a Multimodal Digital Twin, and a State-Updating Digital Twin. The findings indicate that architectural complexity alone did not improve prediction, but different model formulations offered distinct benefits in calibration, explainability, and longitudinal representation.

### 5.1 Principal Findings

The first contribution transformed heterogeneous T1D records into an anchor-based prediction problem. This representation preserved key characteristics of real-world CGM data, such as irregular sampling, incomplete modality availability, monitoring-density variation, and outcome imbalance. Patient-level partitioning was crucial because adjacent anchors from the same patient might contain overlapping temporal information. Although the dataset included over 5.5 million anchors, these represented repeated observations from 736 individuals rather than independent samples.

The second finding was that multimodal and state-updating formulations did not consistently surpass the strong sequence baseline. The digital-twin models achieved several numerical advantages for shorter prediction horizons, while the Hybrid LSTM–Transformer was most effective for both 120-minute outcomes. However, none of the differences remained significant after Holm correction. Consequently, the digital-twin models should be viewed as competitive unified alternatives rather than universally superior predictors.

The third finding was that reliability-oriented evaluation revealed differences not captured by discrimination metrics alone. The Multimodal Digital Twin achieved the lowest mean Brier score, the State-Updating Digital Twin achieved the lowest mean ECE-10, and the Hybrid LSTM–Transformer demonstrated the strongest temporal explanation consistency under its model-specific evaluation protocol. These results underscore that model selection in healthcare computing requires consideration of calibration, interpretability, and behavioral robustness besides accuracy.

### 5.2 Prediction Horizon and Model Complexity

Prediction horizon was a more significant determinant of performance than model architecture. All models exhibited decreasing AUROC, AUPRC, and F1-score from 30 to 120 minutes, with the largest degradation observed for hypoglycemia. This behavior reflects the increasing uncertainty of longer-term glucose evolution, where recent glucose trajectories become progressively insufficient because relevant factors such as insulin administration, carbohydrate intake, physical activity, and behavioral changes were not consistently available. The absence of a consistent long-horizon advantage for the digital-twin models suggests that multimodal fusion and latent-state propagation alone cannot overcome missing physiological and behavioral determinants. However, these components may still add value by improving probability estimates, maintaining patient-specific representations, and supporting continuous-monitoring scenarios.

### 5.3 Role of Multimodal Fusion and State Updating

The Multimodal Digital Twin demonstrated that structured patient information can complement glucose trajectories, although recent temporal information remained the dominant contributor. The relatively smaller contribution of biochemical and static variables suggests their value may come from contextual adjustment and refined calibration rather than substantial gains in discrimination. The State-Updating Digital Twin introduced explicit longitudinal state propagation with gap-aware transitions and validity masking. Although this formulation did not consistently improve predictive metrics, it provides a computational representation of evolving patient status relevant to continuous healthcare monitoring systems. The limited predictive improvement may be partly explained by the overlapping retrospective windows, where each anchor already contained substantial recent history. Therefore, the state-updating formulation's primary contribution is architectural: enabling continuous patient-state representation under irregular longitudinal observations rather than guaranteeing improved prediction at every horizon.

### 5.4 Reliability, Explainability, and Behavioral Analysis

The evaluation beyond discrimination revealed important differences among architectures. Calibration performance varied across models and targets, with longer-horizon hypoglycemia remaining the most challenging task. This suggests that high discrimination does not necessarily yield reliable probability estimates, especially for imbalanced clinical events. Threshold analysis further demonstrated that operating points depend on event type and prediction horizon. Hypoglycemia required higher thresholds than hyperglycemia, emphasizing that alert strategies should consider clinical consequences, false-alert burden, and workflow requirements rather than relying on a universal decision threshold. Explanation analysis showed that recent glucose dynamics dominated model behavior across architectures. The Hybrid LSTM–Transformer exhibited the strongest temporal attribution consistency, while the Multimodal Digital Twin provided stable modality-level interpretation. The State-Updating Digital Twin showed lower single-anchor perturbation effects, which may reflect distributed information encoding across multiple historical states rather than poor interpretability. Future explanation methods for longitudinal models should therefore move beyond single-component perturbations toward trajectory-level analysis. Cross-model disagreement increased with prediction horizon, indicating greater uncertainty in longer-range predictions. Such disagreement may provide additional information for future monitoring systems, where uncertain predictions could trigger conservative handling or human review rather than automatic action.

### 5.5 Implications for Healthcare Computing

This work's main contribution is not a single optimal architecture but a structured evaluation framework for longitudinal healthcare prediction systems. The proposed approach separates predictive performance from deployment-oriented properties, including calibration, explanation reliability, disagreement behavior, and perturbation sensitivity. For healthcare monitoring applications, the State-Updating Digital Twin offers a useful formulation when continuous maintenance of patient state and multimodal integration are required. The Multimodal Digital Twin may be preferable when probability accuracy and modality-level interpretation are prioritized, whereas sequence models remain attractive when simplicity and temporal interpretability are more important. Importantly, the proposed models should be considered decision-support components rather than autonomous clinical systems. Before real-world adoption, prospective validation, workflow integration, recalibration, and human-factor evaluation are required.

### 5.6 Limitations and Future Work

This study has several limitations.

- **Single-dataset evaluation:** The evaluation was conducted on a single retrospective dataset. Although patient-level validation prevented cross-patient data leakage, external validation across institutions, devices, and heterogeneous populations is still required to establish generalizability.

- **Missing contextual determinants:** Several important factors influencing short-term glucose dynamics—including insulin delivery, carbohydrate intake, physical activity, sleep, and stress—were not consistently available. Integrating these variables may improve predictive performance, particularly at longer forecasting horizons.
- **Temporal correlation among anchors:** The anchor-based formulation produces repeated observations from the same patients. Although patient-level splitting prevents leakage across development and test partitions, anchors within the same development partition remain temporally correlated.
- **Limited statistical power:** Statistical comparisons were based on only five outer folds, which limits the power to detect small performance differences among the models. Therefore, the absence of statistically significant differences should not be interpreted as evidence of model equivalence.
- **Operational rather than physiological digital twins:** The proposed digital twins are data-driven operational representations rather than mechanistic physiological simulators. They estimate evolving glycemic risk but do not reproduce individual metabolic physiology or generate automated treatment recommendations.
- **Future work:** Future research should investigate external validation, richer multimodal integration, uncertainty-aware prediction, selective prediction, adaptive recalibration, and trajectory-level explanation methods. Prospective studies should also evaluate whether these models improve clinical monitoring workflows by reducing alert burden, strengthening risk communication, and supporting safer clinical decision-making.

## 6 Conclusion

This study introduced an anchor-based comparative framework for short-term glycemic-risk prediction using longitudinal glucose trajectories and multimodal patient information. We transformed the T1DiabetesGranada records into over 5.5 million standardized prediction anchors from 736 patients, enabling unified prediction of hypoglycemia and hyperglycemia across 30-, 60-, and 120-minute horizons under a patient-level validation protocol. The Hybrid LSTM–Transformer, Multimodal Digital Twin, and State-Updating Digital Twin showed comparable predictive performance, suggesting that increased architectural complexity alone does not improve discrimination. The digital-twin models offered advantages in specific areas, such as multimodal integration and probabilistic reliability, while the sequence baseline remained competitive, particularly for longer prediction horizons. No statistically significant superiority was observed after Holm correction. Beyond predictive performance, the study showed that reliability-oriented evaluation provides complementary insights for healthcare AI systems. The compared models differed in calibration, explanation behavior, disagreement patterns, and responses to controlled perturbations. The State-Updating Digital Twin was selected as the preferred formulation for further investigation due to its combination of joint multi-horizon prediction, multimodal fusion, gap-aware state propagation, and improved calibration characteristics. This selection does not imply universal superiority or clinical readiness. Overall, this work underscores the importance of evaluating longitudinal healthcare prediction models through a multidimensional perspective that integrates predictive accuracy with calibration, interpretability, robustness, and deployment-oriented analysis. Future research should focus on external validation, richer multimodal integration, prospective workflow evaluation, and assessment of clinical utility before translation into patient- or clinician-facing systems.

# A Supplementary Material

## A.1 Comparison of Related Studies

Table S1 summarizes representative studies on glucose forecasting, glycemic-risk prediction, and diabetes digital-twin modelling. The comparison highlights differences in prediction objectives, input modalities, modelling strategies, and evaluation protocols. Because the studies employ heterogeneous datasets and experimental settings, their reported results should not be interpreted as directly comparable performance estimates.

Table S1. Comparison of representative studies on glycemic prediction and diabetes digital-twin modelling. Differences in datasets, prediction objectives, input modalities, validation protocols, and evaluation criteria limit direct comparison across studies.

| Study | Dataset and Inputs | Prediction Objective / Horizon | Model Approach | Evaluation Strategy | Main Limitation |
|---|---|---|---|---|---|
| Mohebbi et al. [27] | T1D cohort; CGM trajectories | Short-term glucose forecasting | Recurrent neural networks | RMSE and MAE | Glucose-value forecasting without multimodal risk assessment. |
| Li et al. [22] | Diabetes CGM datasets | Short-term glucose prediction | Recurrent neural networks | Forecasting performance | Does not address multi-event risk prediction or reliability analysis. |
| Martinsson et al. [26] | T1D; CGM and meal information | Short-term meal and glucose forecasting | Deep learning temporal models | Forecasting error metrics | Focuses on glucose trajectories rather than unified hypo/hyperglycemia risk. |
| Fox et al. [15] | T1D; CGM-derived features | Hypoglycemia prediction; approximately 40 min | Machine learning ensembles | AUROC, sensitivity, and false alarms | Hypoglycemia-specific prediction without broader multimodal reliability assessment. |
| Shao et al. [38] | T1D populations; CGM signals | Short-term hypoglycemia prediction | Deep learning | Discrimination and external validation | Single-event prediction without multimodal digital-twin comparison. |
| Moon et al. [29] | T1D; CGM trajectories | Personalized short-term glucose prediction | Meta-learning LSTM–Transformer | Accuracy and personalization | Focuses on glucose forecasting rather than reliability-aware risk prediction. |
| Karagoz et al. [19] | CGM glucose datasets | Multi-horizon glucose forecasting | Transformer architectures | Forecasting metrics | Limited multimodal clinical integration and reliability assessment. |
| Prendin et al. [32] | T1D; CGM and digital-twin-generated data | Personalized glucose forecasting | Digital twin augmentation and deep learning | Prediction accuracy | Digital twin primarily supports data augmentation rather than adaptive state representation. |
| Roquemen-Echeverri et al. [35] | T1D; physiological and glucose data | Long-term glucose dynamics replication | Physiologically constrained neural digital twin | Simulation similarity and accuracy | Emphasizes physiological replication rather than short-term risk benchmarking. |
| Tran et al. [41] | T1D; CGM and insulin information | Long-horizon glucose forecasting | Physiological digital twin and Seq2Seq LSTM | Forecasting errors | Does not evaluate multitask glycemic-risk reliability. |
| GlyTwin [3] | T1D; CGM and contextual information | Patient-specific intervention modelling | Personalized counterfactual digital twin | Counterfactual validity | Intervention-oriented rather than predictive reliability benchmarking. |
| Monirzadeh et al. [28] | Diabetes simulation; physiological and clinical variables | Decision-aware diabetes modelling | Simulation-driven digital twin | Simulation-based evaluation | Limited empirical clinical validation. |
| **This study** | **T1DiabetesGranada; 736 patients; CGM, demographic, biochemical, diagnostic, missingness, and temporal-gap features** | **Six-task hypo/hyperglycemia risk prediction; 30, 60, and 120 min** | **Hybrid LSTM–Transformer, Multimodal Digital Twin, and State-Updating Digital Twin** | **Discrimination, calibration, thresholds, explanation reliability, disagreement, and perturbation analysis** | **Retrospective evaluation; prospective validation and clinical deployment remain future work.** |

Table S2. Summary of model architectures and training procedures for the three evaluated neural model families.

| Component | Configuration |
|---|---|
| Loss function | Binary cross-entropy with task-specific class weighting estimated from the corresponding training fold. |
| Optimizer | Adam or AdamW, depending on the model configuration. |
| Checkpoint selection | Best validation checkpoint selected within each outer fold before held-out test evaluation. |
| Threshold selection | Target-specific classification thresholds selected from validation predictions by maximizing F1-score and fixed before test evaluation. |
| Hybrid LSTM–Transformer | Target-specific binary prediction models combining bidirectional LSTM encoding, Transformer attention, dropout regularization, and classification heads. |
| Multimodal Digital Twin | Multitask architecture comprising a temporal recurrent–transformer encoder, static MLP branch, biochemical MLP branch, contextual MLP branch, shared fusion representation, and six task-specific output heads. |
| State-Updating Digital Twin | Multitask longitudinal architecture incorporating multimodal anchor embeddings, gap-dependent latent-state decay, GRU-based state updates, validity masking for padded anchors, and six task-specific output heads. |
| Training protocol | Preprocessing, imputation, normalization, class weighting, model training, checkpoint selection, and threshold optimization performed independently within each patient-level outer fold. |
| Implementation | Python, PyTorch, NumPy, pandas, and scikit-learn. |

## A.2 Supplementary Model Configuration

Table S2 summarizes the principal architectural components and training procedures used in the comparative evaluation. These details complement the methodological description in Section 3. Model development, preprocessing, and hyperparameter selection were conducted within patient-level training and validation partitions to maintain separation from held-out test patients.

## A.3 Additional Results for RQ1: Anchor-Based Multimodal Representation

This subsection provides additional descriptive analyses of the anchor-based dataset, including patient-level modality availability, fold distribution, glucose intermeasurement gaps, and outcome prevalence.

*A.3.1 Multimodal Coverage.* Figure S1 presents patient-level modality availability. Glucose records were available for all 736 patients, whereas biochemical and diagnostic information exhibited lower coverage. Missing modalities were retained through explicit availability indicators rather than complete-case filtering.

*A.3.2 Patient-Level Fold Distribution.* All prediction anchors belonging to the same patient were assigned to a single outer fold. Figure S2 illustrates the distribution of held-out prediction anchors across the five patient-level outer folds.

*A.3.3 Intermeasurement Gap Statistics.* Table S3 summarizes glucose intermeasurement-gap statistics. Although most consecutive measurements followed the nominal 15-minute sampling interval, occasional prolonged gaps produced a highly skewed distribution.

*A.3.4 Outcome Distribution.* Figure S3 shows the prevalence of positive hypoglycemia and hyperglycemia events across prediction horizons. Hypoglycemia was substantially less prevalent than hyperglycemia, highlighting differences in target imbalance across the six prediction tasks.

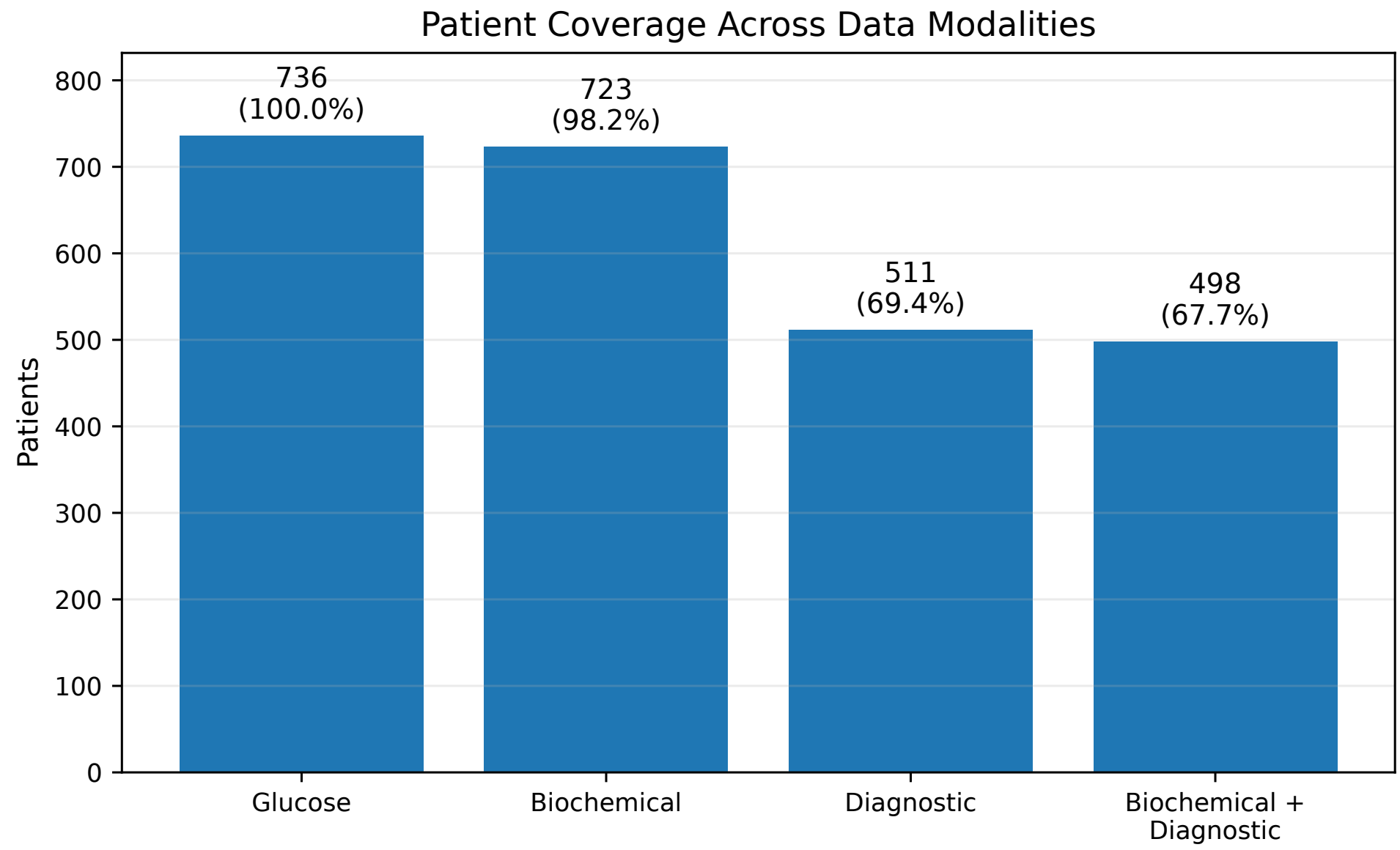

Fig. S1. Patient-level availability of glucose, biochemical, and diagnostic modalities.

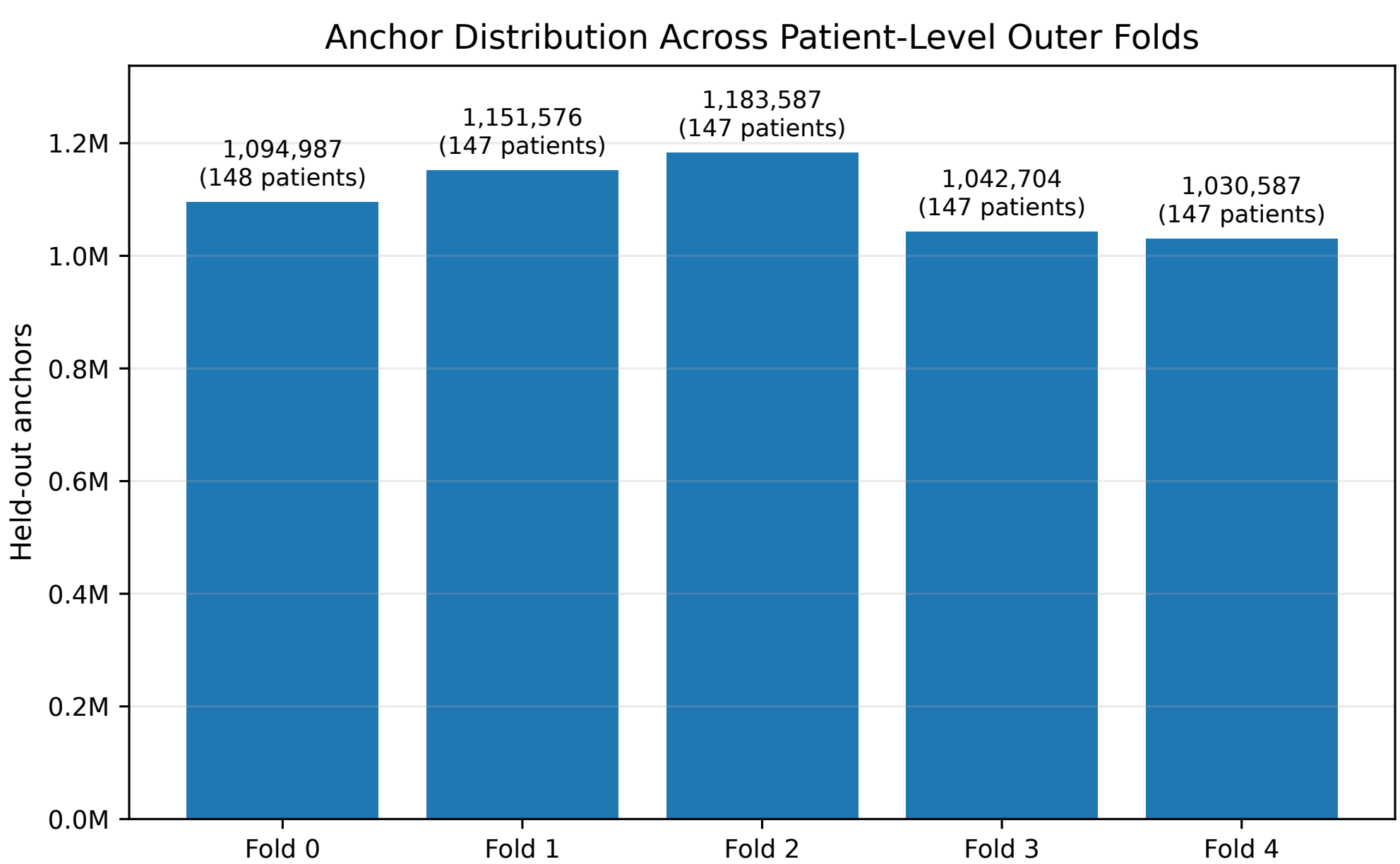

Fig. S2. Held-out anchor distribution across the five patient-level outer folds.

Table S3. Descriptive statistics of glucose intermeasurement gaps.

| Characteristic | Value |
|---|---|
| Number of gaps | 22,648,943 |
| Minimum | 1 min |
| Q1 | 15 min |
| Median | 15 min |
| Q3 | 15 min |
| Mean | 20.76 min |
| Maximum | 1,516,981 min |

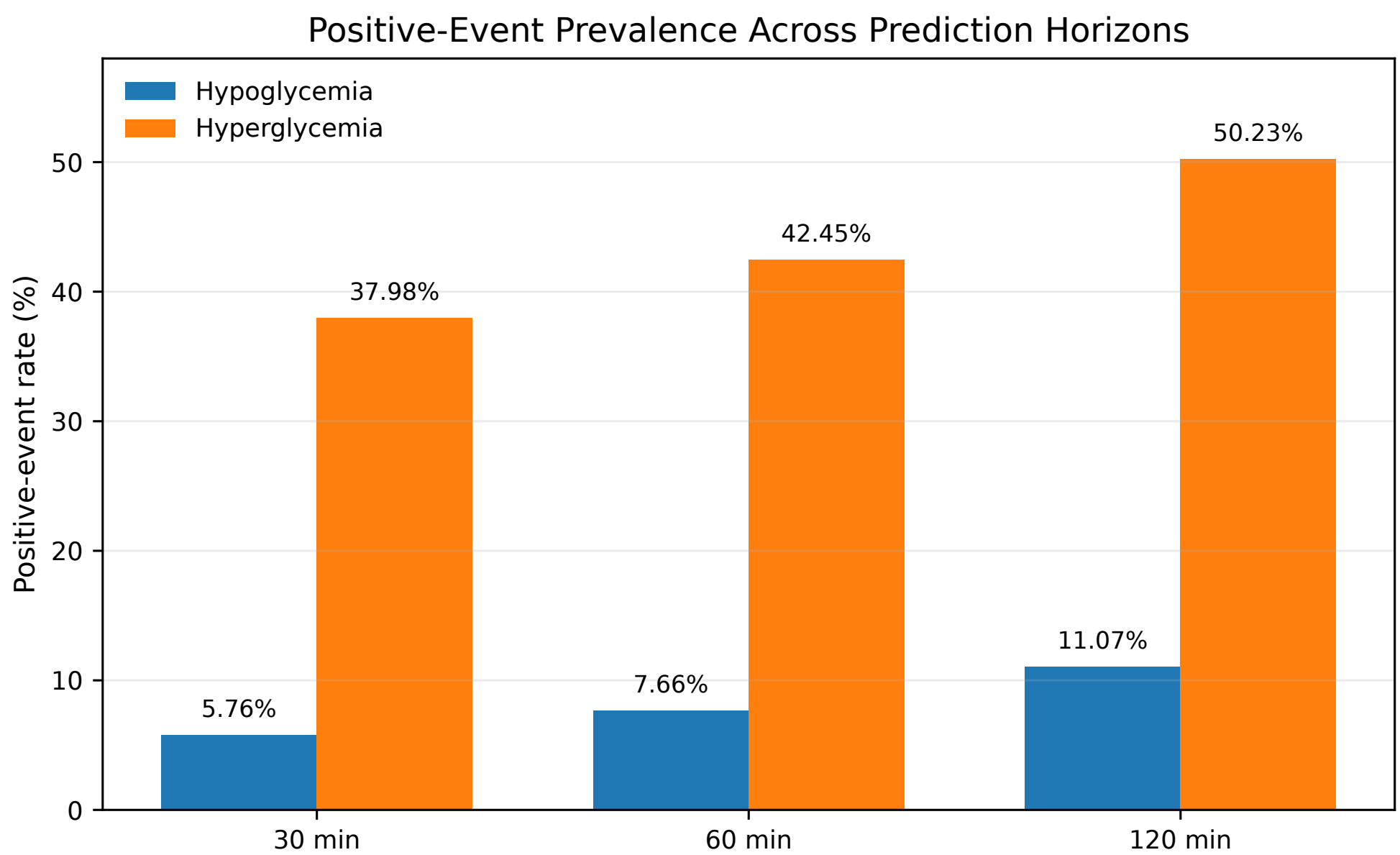

Fig. S3. Positive-event prevalence across hypoglycemia and hyperglycemia prediction horizons.

## A.4 Additional Results for RQ2: Predictive Performance

This subsection presents supplementary visual comparisons of AUPRC, F1-score, and numerical model rankings. Detailed performance estimates and statistical comparisons are reported in the main manuscript.

*A.4.1 AUPRC and F1-Score Comparison.* Figures S4 and S5 summarize mean AUPRC and F1-score across prediction targets. The three model families exhibited broadly similar performance patterns, with predictive performance generally decreasing at longer horizons, particularly for hypoglycemia.

*A.4.2 Numerical Model Rankings.* Figure S6 identifies the model achieving the highest numerical mean performance for each prediction target and evaluation metric. No model family consistently dominated across all tasks. These rankings are descriptive and should not be interpreted as evidence of statistically significant differences.

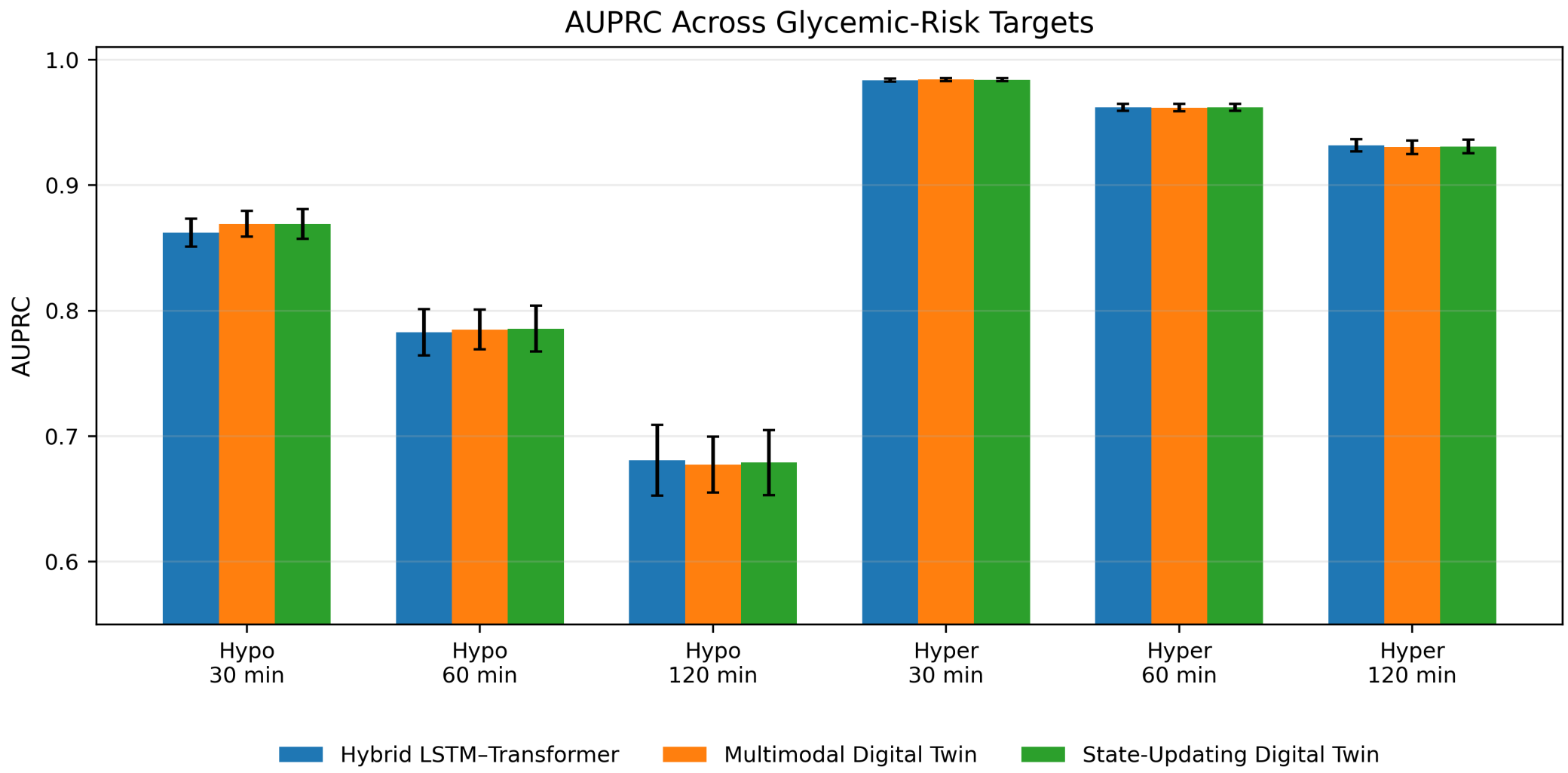

Fig. S4. Mean AUPRC across prediction targets and model families.

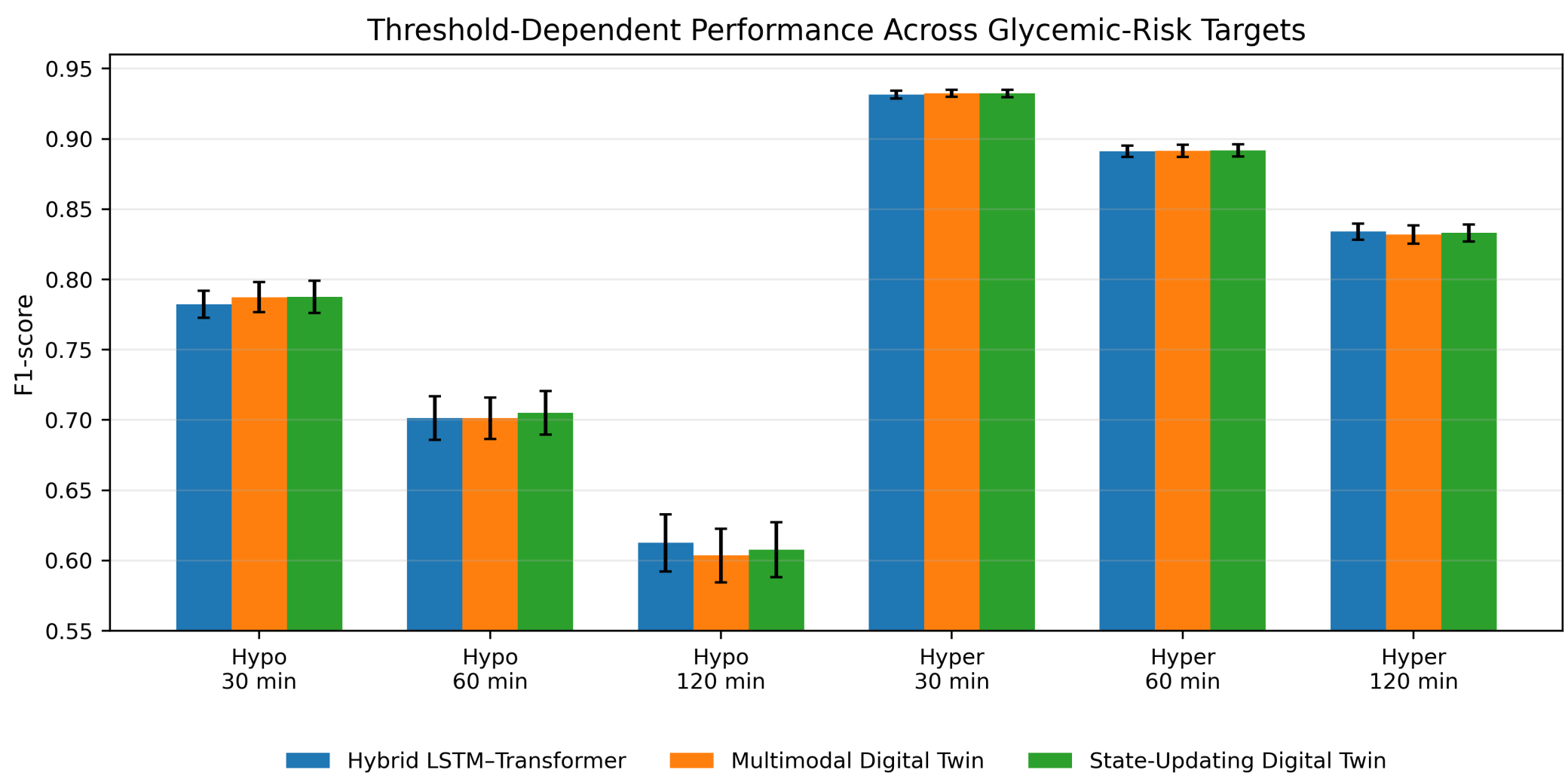

Fig. S5. Mean F1-score across prediction targets and model families.

## A.5 Additional Results for RQ3: Reliability and Explainability

This subsection provides supplementary analyses of local explanations, explanation faithfulness and stability, cross-model disagreement, subgroup errors, and controlled perturbation responses. These analyses complement the predictive evaluation by examining model behaviour beyond discrimination metrics.

*A.5.1 Local Explanation Examples.* Figure S7 presents representative SHAP explanations for selected true-positive, false-positive, false-negative, and true-negative hyperglycemia prediction cases. The examples illustrate case-specific attribution patterns and are intended for qualitative inspection rather than population-level inference.

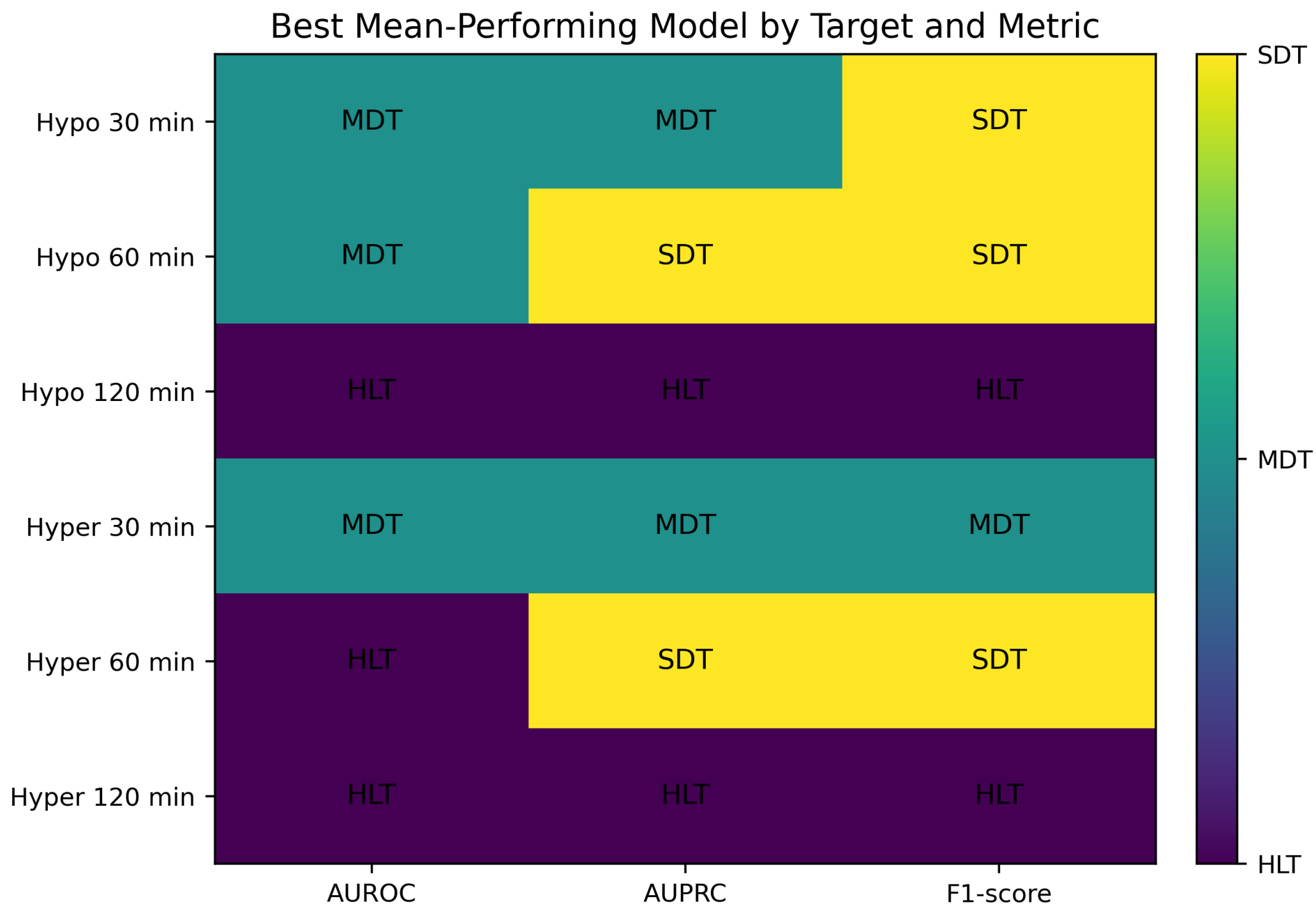

Fig. S6. Numerical model rankings across prediction targets and evaluation metrics. Rankings reflect mean performance and do not establish statistical superiority.

*A.5.2 Explanation Faithfulness and Stability.* Figure S8 presents target-specific explanation faithfulness and stability under the respective model-specific evaluation protocols. Because the evaluated components and perturbation procedures differ across architectures, these values should not be directly ranked across model families.

*A.5.3 Cross-Model Disagreement and Error Stratification.* Figures S9 and S10 present additional descriptive analyses of cross-model disagreement and subgroup error patterns. These analyses characterize variation in model predictions and errors without establishing subgroup-specific clinical validity.

*A.5.4 Recent-Glucose Perturbation Analysis.* Table S4 summarizes model responses to controlled changes in recent glucose observations. The reported values represent ranges of mean probability changes across prediction horizons. These analyses characterize directional model sensitivity and should not be interpreted as validated physiological counterfactual effects.

*A.5.5 State-History Perturbation Analysis.* Figure S11 illustrates the response of the State-Updating Digital Twin to perturbations of its longitudinal state history. This analysis provides additional descriptive evidence regarding the model's sensitivity to historical information.

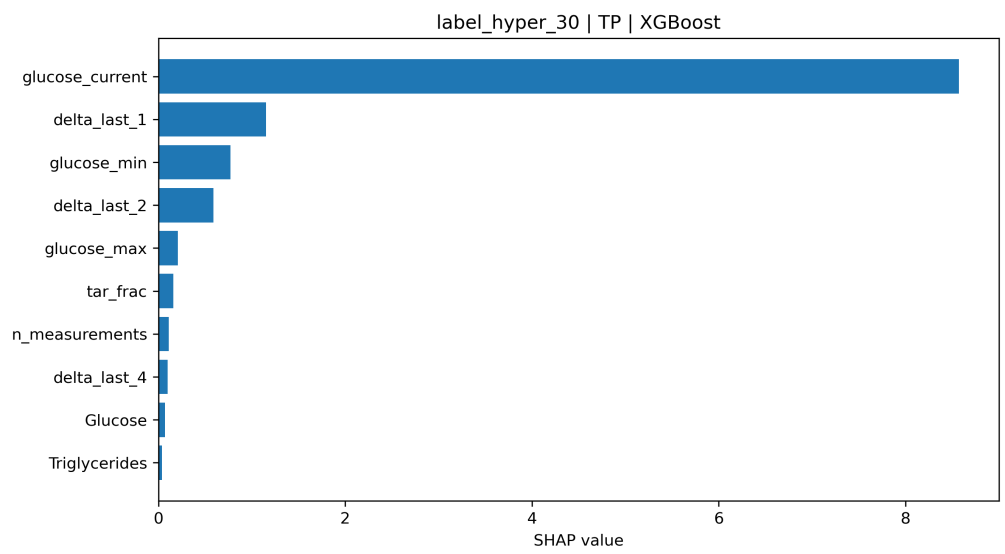

(a) True positive, 30 min.

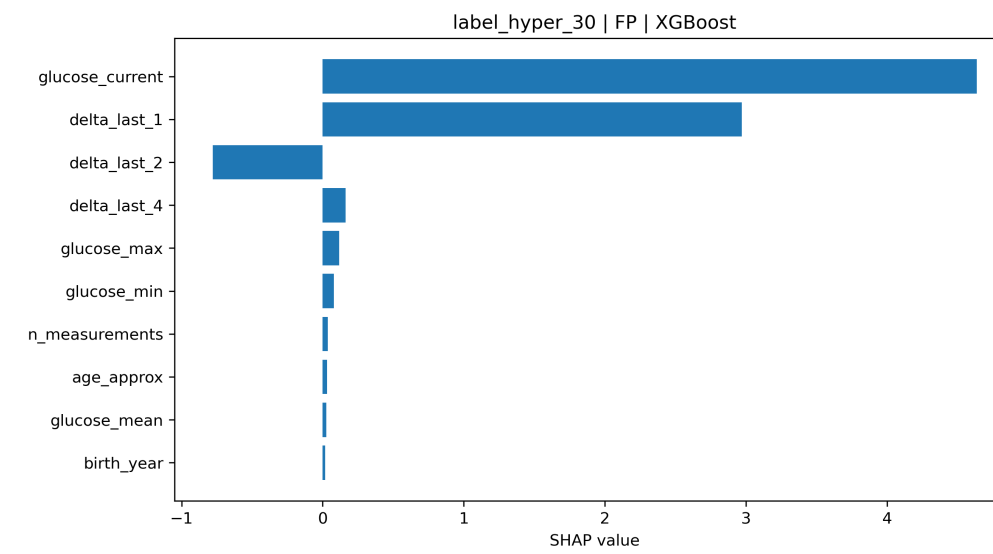

(b) False positive, 30 min.

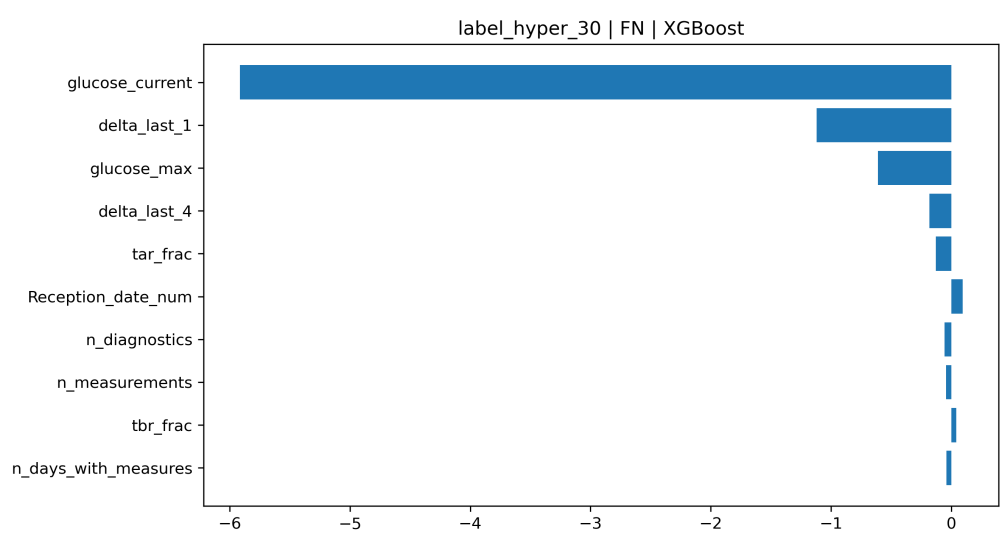

(c) False negative, 30 min.

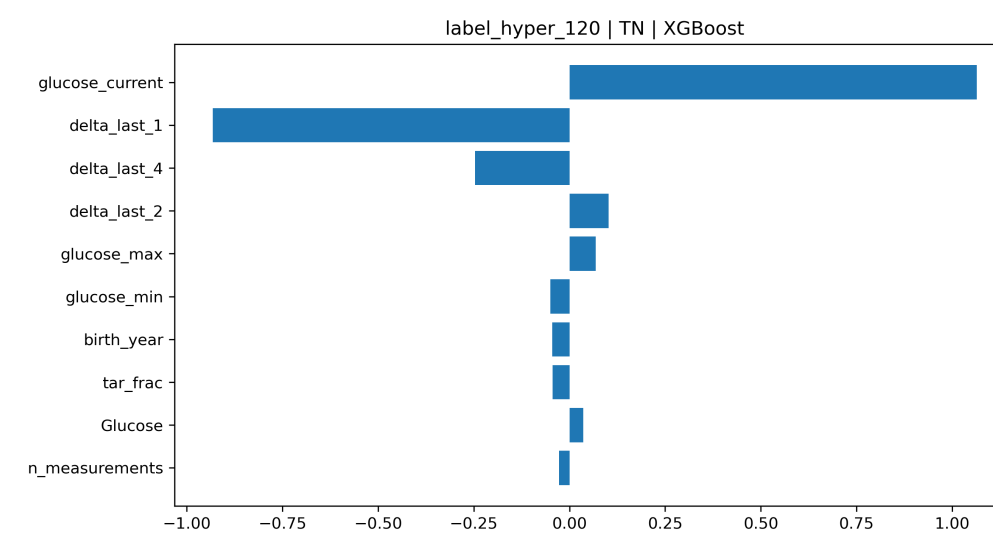

(d) True negative, 120 min.

Fig. S7. Representative SHAP explanations for selected hyperglycemia prediction cases at 30- and 120-minute horizons. The examples illustrate heterogeneous attribution patterns and are provided for qualitative inspection only.

Table S4. Recent-glucose perturbation responses across the principal model families. Values represent ranges of mean probability changes across prediction horizons.

| **Model family** | **Hypoglycemia perturbation** | **Mean effect range** | **Hyperglycemia perturbation** | **Mean effect range** |
|---|---|---|---|---|
| Hybrid LSTM–Transformer | Lower recent glucose | 0.401–0.466 | Higher recent glucose | 0.417–0.445 |
| Multimodal Digital Twin | Lower recent glucose | 0.661–0.818 | Higher recent glucose | 0.484–0.641 |
| State-Updating Digital Twin | Lower recent glucose | 0.583–0.751 | Higher recent glucose | 0.480–0.586 |

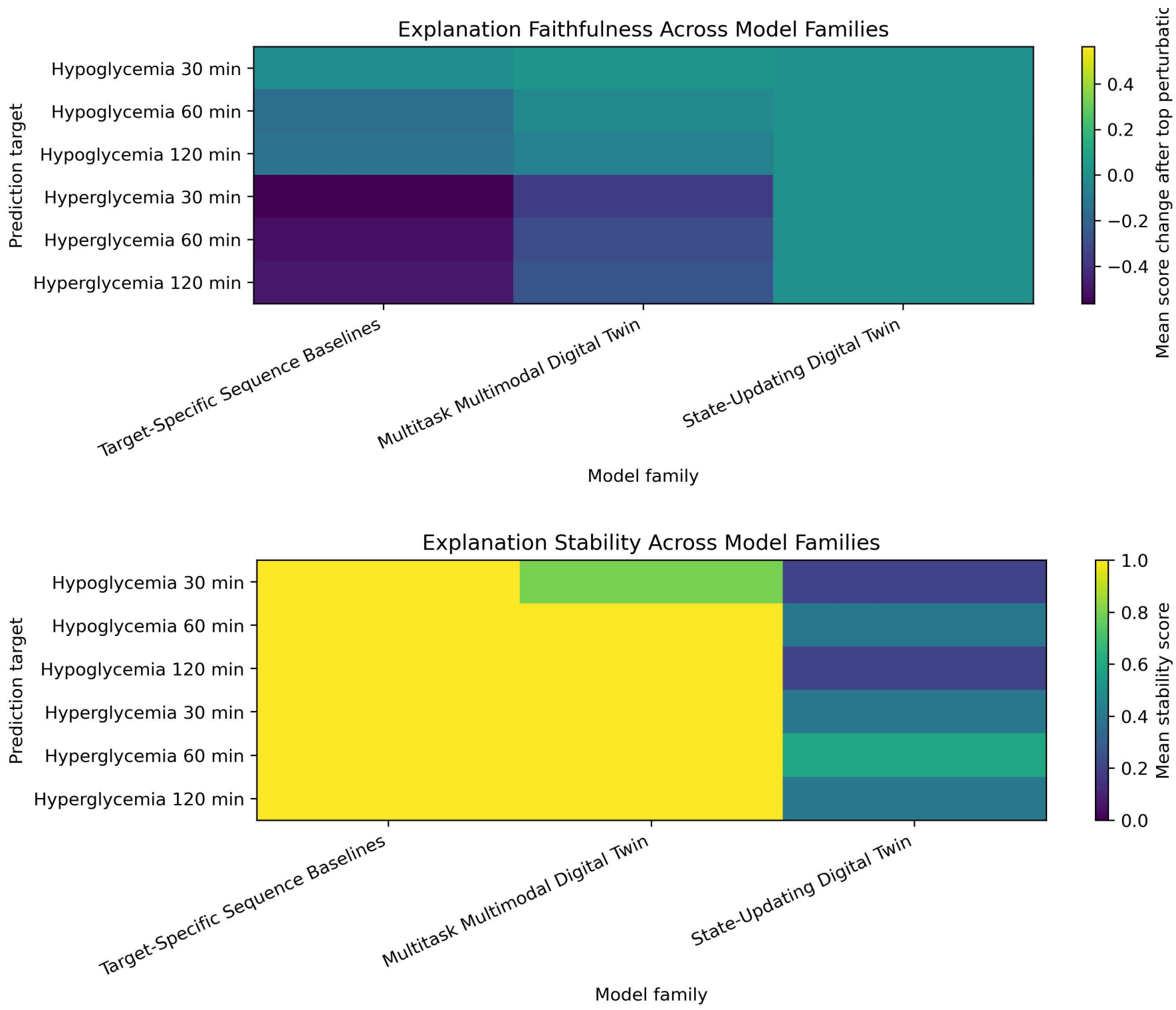

Fig. S8. Target-specific explanation faithfulness and stability under model-specific evaluation protocols. Results are descriptive and are not directly comparable across architectures.

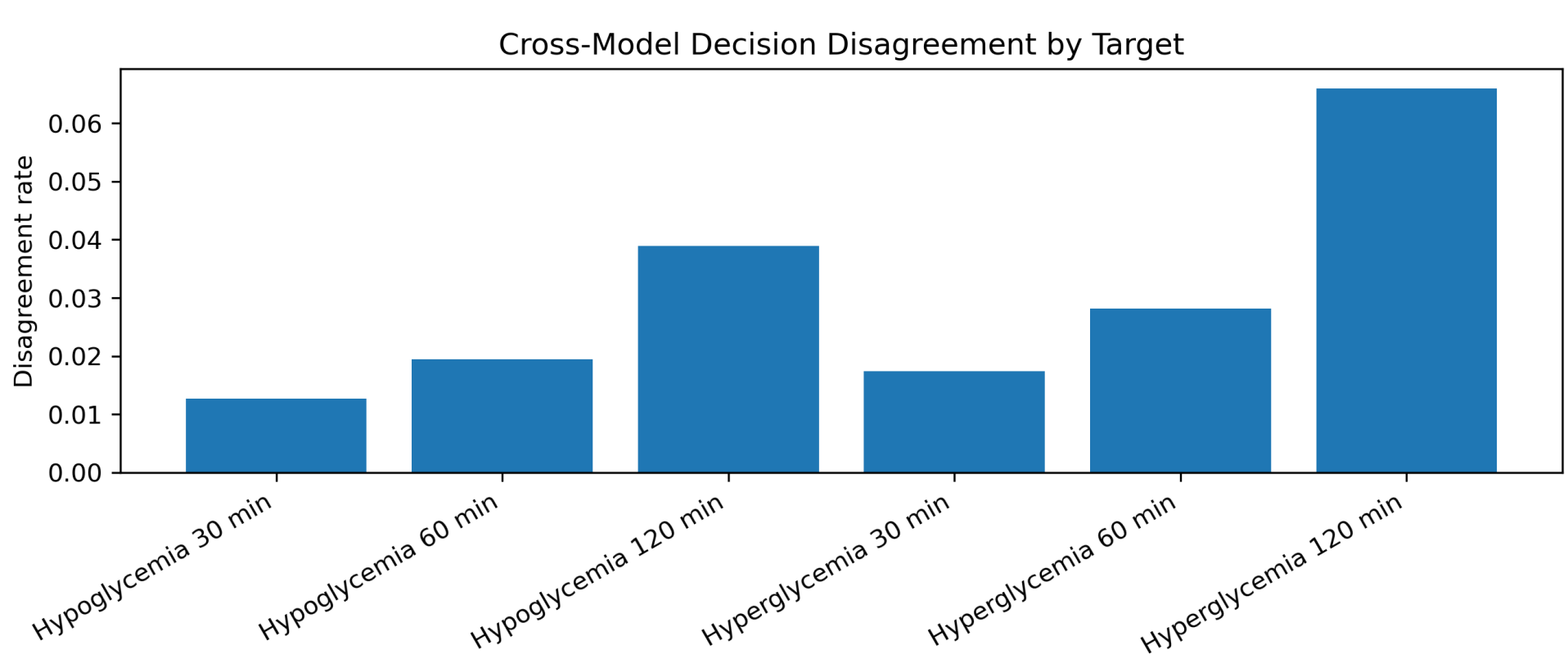

Fig. S9. Cross-model disagreement across hypoglycemia and hyperglycemia prediction targets.

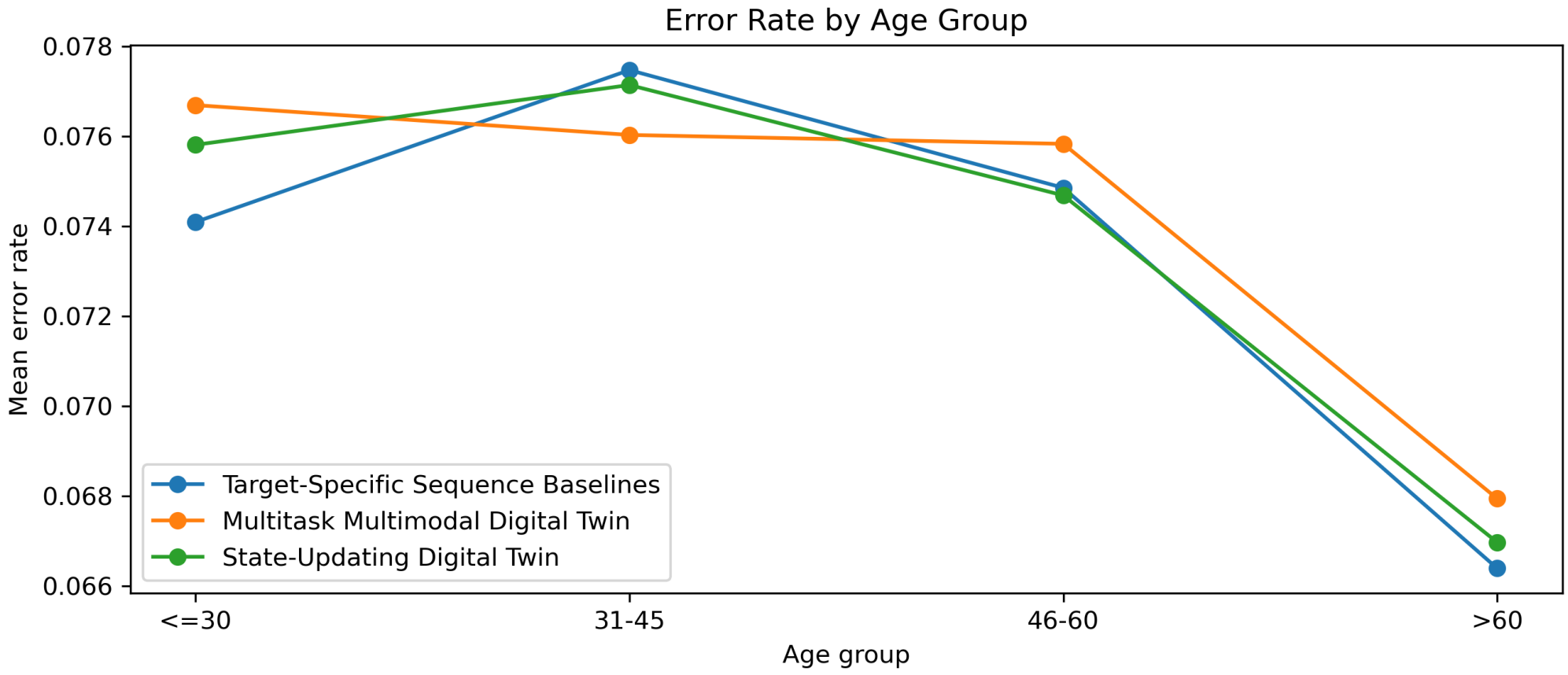

Fig. S10. Descriptive subgroup error analysis by age group.

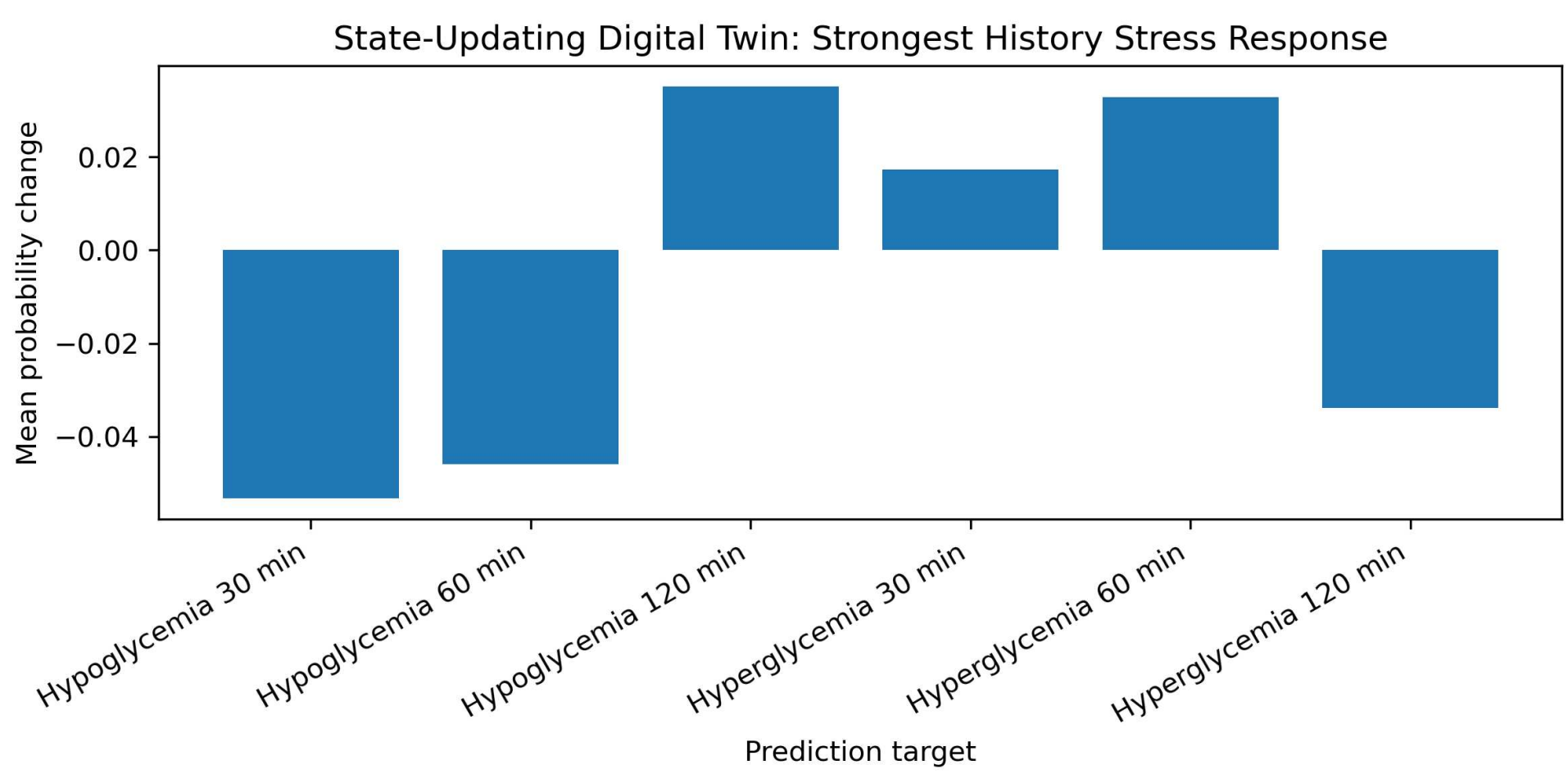

Fig. S11. State-history perturbation response for the State-Updating Digital Twin.